\documentclass{article}

\usepackage{arxiv}
\usepackage{amsmath}
\usepackage{multirow}
\usepackage{mathrsfs}
\usepackage{graphicx}
\usepackage{balance}
\usepackage[utf8]{inputenc} 
\usepackage[T1]{fontenc}    
\usepackage{hyperref}       
\usepackage{url}            
\usepackage{booktabs}       
\usepackage{amsfonts}       
\usepackage{nicefrac}       
\usepackage{microtype}      
\usepackage{cleveref}       
\usepackage{lipsum}         
\usepackage{graphicx}
\usepackage{natbib}
\usepackage{doi}

\title{Topology-Aware Latent Diffusion for 3D Shape Generation}

\newif\ifuniqueAffiliation

\author{Jiangbei Hu$^{*,1,2}$, Ben Fei$^{*,3}$, Baixin Xu$^{1}$, Fei Hou$^{4}$, Weidong Yang{$^{\dagger,3}$}, Shengfa Wang$^{2}$, Na Lei$^{2}$, \\\textbf{Chen Qian}$^{5}$, \textbf{Ying He}{$^{\dagger, 1}$}\\
	$^1$Nanyang Technological University,\quad$^2$Dalian University of Technology,\\
	 $^3$Fudan University,\quad $^4$Chinese Academy of Sciences,\quad $^5$SenseTime Research\\
	\texttt{jiangbei.hu@ntu.edu.sg},\quad \texttt{bfei21@m.fudan.edu.cn},\quad \texttt{yhe@ntu.edu.sg}\\
}
 
\renewcommand{\shorttitle}{\textit{arXiv} Template}

\hypersetup{
pdftitle={A template for the arxiv style},
pdfsubject={q-bio.NC, q-bio.QM},
pdfauthor={David S.~Hippocampus, Elias D.~Striatum},
pdfkeywords={First keyword, Second keyword, More},
}

\begin{document}
\maketitle

\newcommand\blfootnote[1]{%
\begingroup
\renewcommand\thefootnote{}\footnote{#1}%
\addtocounter{footnote}{-1}%
\endgroup
}
\blfootnote{{$*$}Equal contribution, {$\dagger$}Corresponding author.}

\begin{abstract}
	We introduce a new generative model that combines latent diffusion with persistent homology to create 3D shapes with high diversity, with a special emphasis on their topological characteristics. Our method involves representing 3D shapes as implicit fields, then employing persistent homology to extract topological features, including Betti numbers and persistence diagrams. The shape generation process consists of two steps. Initially, we employ a transformer-based autoencoding module to embed the implicit representation of each 3D shape into a set of latent vectors. Subsequently, we navigate through the learned latent space via a diffusion model. By strategically incorporating topological features into the diffusion process, our generative module is able to produce a richer variety of 3D shapes with different topological structures. Furthermore, our framework is flexible, supporting generation tasks constrained by a variety of inputs, including sparse and partial point clouds, as well as sketches. By modifying the persistence diagrams, we can alter the topology of the shapes generated from these input modalities.
\end{abstract}

\keywords{Generative model \and Persistent homology \and Latent diffusion model \and Topology-aware generation}

\section{Introduction}
In the realms of computer graphics and 3D vision, the automatic creation and reconstruction of 3D shapes via deep learning techniques have emerged as a significant area of interest~\cite{wang2021neus,yariv2021volume,xu2023survey}. Methods based on artificial intelligence have the potential to streamline the design and modeling process of 3D shapes, while concurrently expanding the availability of feature-enriched 3D models.
These methods employ generative networks such as Variational Autoencoders (VAEs) and Generative Adversarial Networks (GANs) to understand the underlying probability distribution within a given 3D shape dataset. They then generate novel shapes by sampling from this learned distribution~\cite{groueix2018papier,zheng2022sdf,chen2021decor}. Furthermore, the incorporation of a conditional mechanism within these generative models facilitates the creation of 3D shapes from diverse inputs, encompassing point clouds, images, and text.
Recently, diffusion models have achieved remarkable success in the domain of image generation~\cite{rombach2022high}. This breakthrough has inspired the development of numerous 3D diffusion models~\cite{xu2023dream3d,luo2021diffusion,zhou20213d}, which have significantly improved both the quality and diversity of the resultant shapes.

Most existing 3D generative networks learn the prior distribution by extracting geometric local features through a conventional encoder, such as CNN or MLP. However, global topological features, which are crucial for the aesthetic and functional attributes of 3D shapes, are often overlooked.
The importance of topological features in 3D shape generation and modeling is particularly manifest in two aspects.
First, by integrating topological features into the learning paradigms of generative models, the models are guided to learn distributions that are influenced not solely by geometric details but also by topological characteristics, resulting in structures that exhibit enhanced 3D complexity.
Moreover, the generation of 3D shapes from sparse point clouds or images often encounters a deficiency in explicit topological information, leading to end products that lack the intended structural integrity. Consequently, it is crucial to incorporate constraints on topological properties during the generation process.
To effectively integrate topological features within generative networks, the primary challenge resides in devising appropriate representations for these features.
Structure-based methods, such as PQ-Net~\cite{wu2020pq} and DSG-Net~\cite{yang2022dsg}, strive to construct 3D objects with complex topological structures by segmenting shapes into fundamental primitives and discerning the connections among different segments. Nevertheless, these methods are intricate because of the requirement for segmentation and reassembly, and they may not be suitable for general objects that cannot be divided into primitives.

In this work, we introduce a new topology-aware generative model that combines the latent diffusion model with persistent homology. Distinct from structure-based methods, our approach harnesses persistent homology (PH) analysis to extract global topological features from 3D shapes. These features then steer the shape generation process within our model.
Specifically, our model first employs a filtration process on the cubical complexes, predicated on the signed distance values associated with the 3D shapes. This procedure yields Betti numbers and persistence diagrams (PDs), which serve as quantitative reflectors of the structure's multi-scale topological characteristics.
Subsequently, we embed the 3D shapes into a lower-dimensional latent space, where we train the diffusion process to produce novel shape variations. Inspired by the 3DShape2VecSet~\cite{zhang20233dshape2vecset}, we train an AutoEncoder based on the transformer architecture to represent each shape as a set of latent vectors. After adding noise to the latent vectors, the network undergoes training for denoising, a step that is pivotal for learning the distribution of 3D shapes via latent space diffusion. Furthermore, our model integrates a conditioning mechanism within the denoising block, leveraging topological features to direct the shape generation.
By specifying different topological features, our methodology enhances the diversity of 3D shapes generated by the diffusion model. In addition, we can generate shapes from other joint conditional inputs, such as sparse/partial point clouds and sketches. Within our framework, the topology of the generated shapes can be adjusted by modifying the corresponding PDs.
We demonstrate the effectiveness of our approach in various shapes from the ShapeNet dataset~\cite{shapenet2015} and the ABC dataset~\cite{Koch_2019_CVPR}.

We summarize the contributions as follows:
\begin{itemize}
	\item Our proposed novel framework is capable of generating 3D shapes with enhanced diversity by utilizing topological features.
	\item We can create 3D models with the desired topology from various inputs that may exhibit topological ambiguity, such as sparse/partial point clouds, as well as sketches.
	\item We can alter the topology of the generated shapes by manipulating the persistence diagrams.
\end{itemize}

\begin{figure*}[!t]
	\centering
	\includegraphics[width=1.0\linewidth]{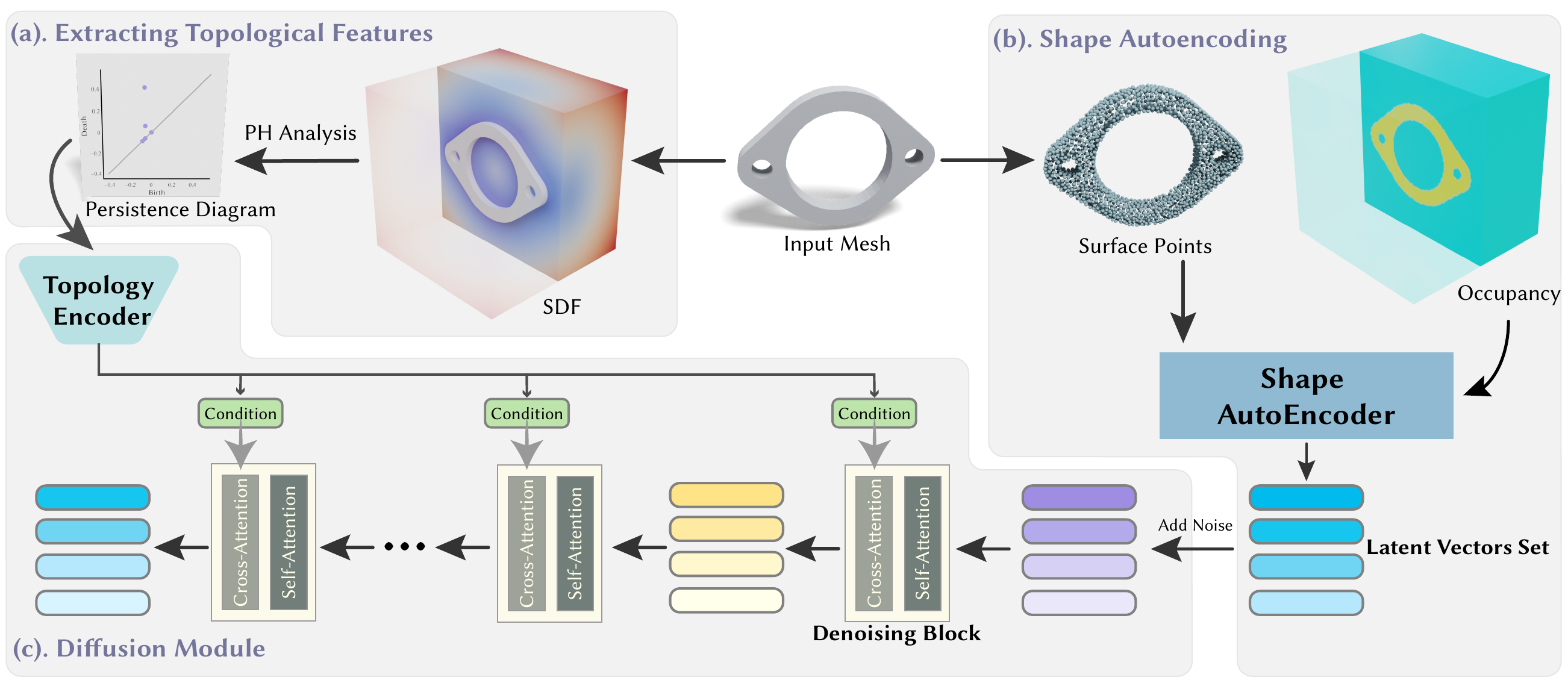}
	\caption{Our framework primarily comprises three steps: (a). The process starts with a persistent homology (PH) analysis applied to the signed distance field (SDF) of the input mesh. This analysis yields topological features, which are represented by persistence diagrams (PDs). (b). Subsequently, dense point clouds sampled on the mesh surface are input to train a shape AutoEncoder. This training phase is supervised by the occupancy field, resulting in a set of latent vectors that represent the 3D shapes. (c). The final step involves training the diffusion model within the previously learned latent space. The diffusion process is made aware of the topology by conditioning each denoising block with features derived from the PDs.}
	\label{fig:pipeline}
\end{figure*}

\section{Related Work}
\label{sec:related}

\subsection{Diffusion Models for 3D Shape Generation.}
Diffusion models are likelihood-based generative models~\cite{croitoru2023diffusion}, consisting of a diffusion process and a reverse process.
These methods have been extensively investigated in the context of image generation~\cite{ho2020denoising, dhariwal2021diffusion, nichol2021improved, ho2022cascaded} and speech synthesis~\cite{kong2020diffwave, popov2021grad}.
To address the issue of slow sampling speed in diffusion models, latent diffusion models~\cite{rombach2022high, vahdat2021score} are proposed to train diffusion models in the latent space of an AutoEncoder, where input data samples are encoded into a more compact representation. This approach significantly improves the training and sampling speed of diffusion models. Consequently, our method builds upon the foundations of latent diffusion models.
The application of diffusion models in 3D generation has also seen a surge of work. These models are initially employed in the generation of point clouds~\cite{luo2021diffusion, zhou20213d, zeng2022lion, lyu2023controllable, nichol2022point}.
Some methods~\cite{zeng2022lion, lyu2023controllable} propose the reconstruction of meshes from the generated point clouds using surface reconstruction techniques~\cite{peng2021shape}.
Other works~\cite{li2023diffusion, chou2023diffusion, shue20233d, liu2023meshdiffusion, zheng2023locally, gupta20233dgen, nam20223d} utilizes implicit fields~\cite{park2019deepsdf, mescheder2019occupancy} to generate meshes. These studies typically employ latent representations such as points, voxels, or triplanes to represent the implicit fields and subsequently train diffusion models on these latent representations. However, few existing works have explicitly considered the topological features of 3D structures in the diffusion generation process, which are significant for the appearance and function of 3D objects.

\subsection{Topology-aware shape modeling}
Topological invariant, such as Betti numbers, Euler characteristics, and genus, play a crucial role in guaranteeing the accuracy of geometric design and reconstruction tasks. \citet{sharf2006competing} developed an automated method based on genus control to accurately reconstruct a surface from a point cloud with the correct topology.
\citet{huang2017topology} presented the first technique for reconstructing multi-labeled material interfaces, providing explicit control over the number of connected components and genus. However, classical topological invariants are basic quantities that are sensitive to topological perturbations and difficult to relate to the geometric scale of topological structures.

Persistent homology (PH)~\cite{edelsbrunner2008persistent} is a powerful method used in topological data analysis (TDA) to capture and quantify multi-scale topological features of the underlying structure, e.g., connected components, loops, and voids. Recent advancements in this field have seen the development of innovative methods that leverage PH analysis for 3D modeling. \cite{zhou2022learning} introduced TopologyNet, an end-to-end network capable of fitting topological representations directly from input point cloud data. For geometric reconstruction, \cite{bruel2020topology} presented a methodology that integrates topological priors into surface reconstruction processes from point clouds. They proposed a topology-aware likelihood function that uses PH measures to preserve or create the desired topology.
\cite{mezghanni2021physically} utilized topological loss from the persistence diagram to ensure the stability and connectivity of the generated structures.
\cite{dong2022topology} further proposed a topology-controllable implicit surface reconstruction method, which optimizes the control coefficients of B-spline functions through a topological target function defined by persistent pairs in persistence diagrams.

\section{Persistent Homology for 3D Shapes}
\label{sec:ph}
The persistent homology (PH) method involves several key steps: constructing a filtration of simplicial complexes, computing the homology groups for these complexes, and performing persistence analysis. In our approach, we employ the cubical complex and the signed distance field (SDF) to conduct PH analysis for 3D shapes.

\begin{figure}[htbp]
	\centering
	\includegraphics[width=0.85\linewidth]{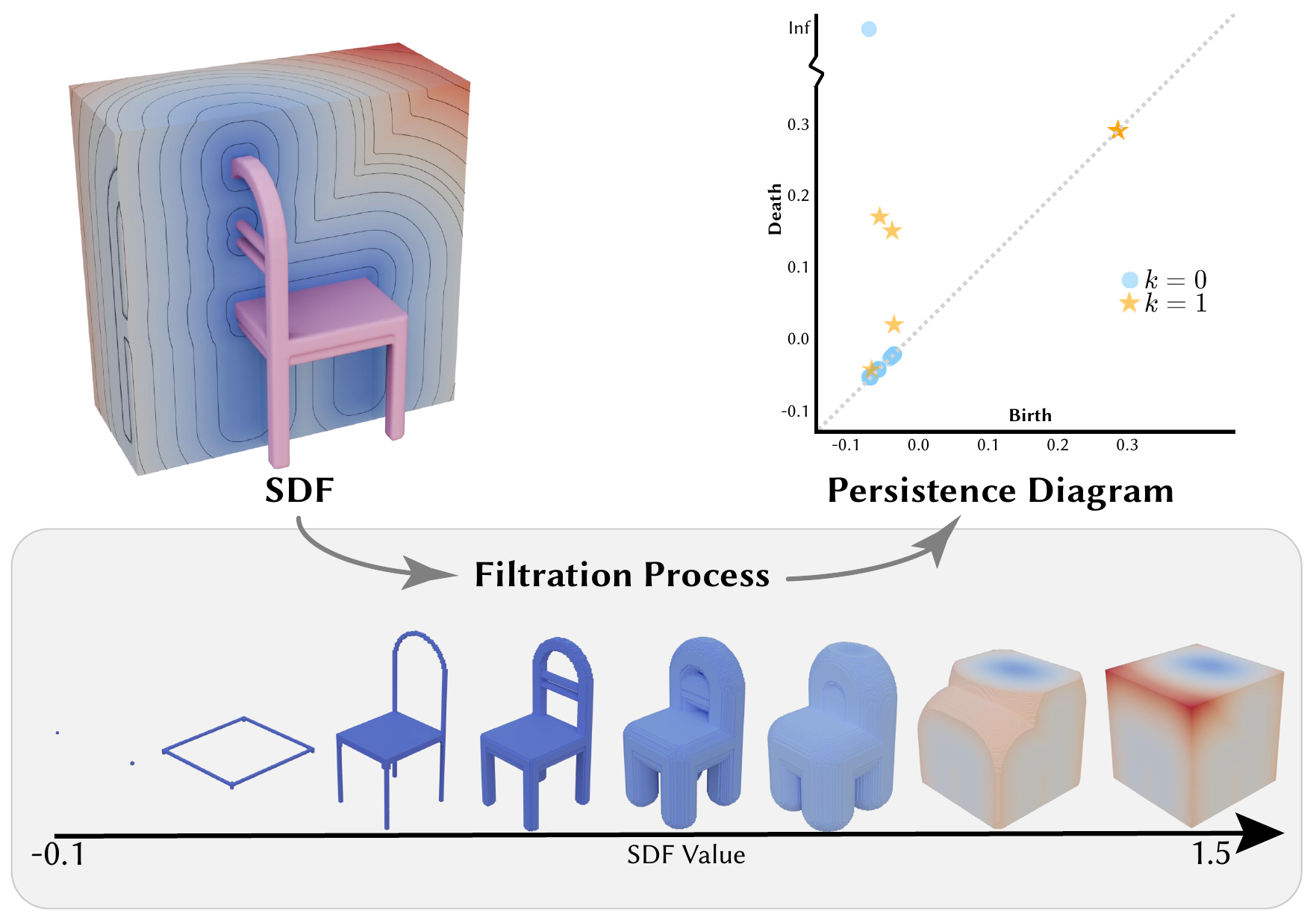}
	\caption{Taking the implicit representations of 3D shapes (SDFs) as filtration values, we utilize the cubical complex to perform the persistent homology analysis for the shapes.
		The multi-scale topological features can be depicted by the persistence diagrams, where each point indicates the birth value and death value of the corresponding topological feature, respectively.}
	\label{fig:cubical}
\end{figure}

\paragraph{Cubical complex.}
PH is capable of dealing with various data types, including meshes~\cite{scaramuccia2020computing}, point clouds~\cite{de2022ripsnet}, and tensors like images~\cite{clough2020topological}. Each data type has a suitable coreponding complex structure used for analyzing~\cite{otter2017roadmap}, such as the \v{C}ech complex and the Vietoris-Rips complex.
Given that we use implicit fields to represent 3D shapes, the use of the cubical complex is a natural choice~\cite{allili2001cubical,wagner2011efficient}.
Let $c^n=I_1\times I_2\times\cdots\times I_n\subset \mathbb{R}^n$ denote the $n$-dimensional elementary cube, where $I=[0, 1]$ or $I=[0, 0]$ (degenerate to a point) is the unit interval. Intuitively, the elementary cubes of a 3D grid include vertices ($c^0$), edges ($c^1$), squares ($c^2$), and voxels ($c^3$).
A \textbf{cubical complex} $\mathcal{K}$ is the union of a set of elementary cubes that satisfy: 1) if a cube $c \in \mathcal{K}$, then all faces of $c$ are also in $\mathcal{K}$; 2) if two cubes $c_1$ and $c_2$ intersect, then their intersection $c_1 \cap c_2$ is a common face of $c_1$ and $c_2$. As shown in Fig.~\ref{fig:cubical}, we can directly construct the cubical complex from the implicit fields of 3D shapes to perform the PH analysis.

\paragraph{Homology group.}
For the $d$-th chain group $\mathbb{C}_d(\mathcal{K})$ of a cubical complex $\mathcal{K}$, we define the $d$-th cycle group $\mathbb{Z}_d=\text{Ker} \partial_d=\{\sigma\in \mathbb{C}_d \mid \partial_d \sigma=0\}$ and the $d$-th boundary group $\mathbb{B}_d=\text{Im} \partial_{d+1}=\{\sigma\in \mathbb{C}_d \mid \exists \hat{\sigma}\in \mathbb{C}_{d+1},\, s.t.,\, \sigma=\partial_{d+1} \hat{\sigma}\}$, where $\partial_d: \mathbb{C}_d\rightarrow \mathbb{C}_{d-1}$ is the boundary operator satisfying $\partial_0=1$ and $\partial_{k-1}\partial_k=0$. The $d$-th \textbf{homology group} is the quotient group $\mathbb{H}_d=\mathbb{Z}_d / \mathbb{B}_d$. The \textbf{Betti number} $\beta_d$ is the rank of $\mathbb{H}_d$ as $\beta_d=\text{rank} \mathbb({H}_d)=\text{rank} (\mathbb{Z}_d) - \text{rank} (\mathbb{B}_d)$, which indicates discrete topological features, such as the number of connected components ($\beta_0$), loops ($\beta_1$), and voids ($\beta_2$).

\paragraph{Filtration process.}
The key concept of PH analysis is the filtration process, which generates a sequence of topological spaces at different scales. As illustrated in Fig.~\ref{fig:cubical}, we grid the domain containing the 3D model to obtain the complete cubical complex $\mathcal{K}$. Subsequently, we define a series of filtered complexes based on the SDF values $S(\sigma)$ assigned to each complex. Given a threshold $s_i$, we define $\mathcal{K}^i=\{\sigma\in\mathcal{K} \mid S(\sigma)\leq s_i \}$. Therefore, by gradually increasing the filtration value, we can obtain a nested sub-complexes sequence $\emptyset\subset\mathcal{K}^0\subseteq\mathcal{K}^1\subseteq\cdots\subseteq\mathcal{K}^N=\mathcal{K}$. Then we represent the $p-$persistent  $d$-th homology group at filtration value $s_i$ as $\mathbb{H}_d^{i,p}=\mathbb{Z}_d^i/(\mathbb{B}_d^{i+p}\bigcap \mathbb{Z}_d^i)$, where $\mathbb{Z}_d^i=\mathbb{Z}_d(\mathcal{K}^i)$, $\mathbb{B}_d^i=\mathbb{B}_d(\mathcal{K}^i)$. The $p$-persistent $d$-th Betti number $\beta_d^{i,p}$ of $\mathcal{K}^i$ is the rank of $\mathbb{H}_d^{l,p}$. As a result, the sequence of nested filtered complexes can capture the construction (birth) and destruction (death) of topological features with different dimensions. For the $l$-th $n$-dimensional topological feature, it corresponds to a pair $(b_l^n, d_l^n)$, where $b_l^n$ and $d_l^n$ are the birth and death filtration values, respectively. The magnitude $|d_l^n-b_l^n|$ indicates the persistence of the topological feature, providing a geometric measure of the topological invariant. Topological characteristics can be encapsulated within a \textbf{persistence diagram (PD)} $\mathcal{D}$, which plots these birth-death pairs in $\mathbb{R}^2$. Analysis of the PD facilitates the discernment of the scale or persistence of topological features; points closer to the diagonal indicate shorter existing time of the corresponding topological features in the homology sequence, and they are more likely to be identified as noise points.

Compared to other complexes derived from point clouds or meshes (e.g. Vietoris-Rips complexes), cubical complexes constructed from SDFs are better suited for shape modeling frameworks that utilize implicit neural representations. SDF serves as a continuous function, offering a smoother filtration process for PH analysis and consequently yielding less noise. Additionally, the persistence of a particular topological feature inherently captures the variations between different levels of implicit fields. This enables a direct correlation with 3D modeling, making it possible to alter the topology of generated shapes by modifying the persistence diagrams.

\section{Topology-aware Latent Diffusion}
We represent 3D shapes using implicit fields and then embed these fields into a lower-dimensional latent space via an auto-encoding module (Sec.~\ref{sec:3d-rep}). Then, we apply PH analysis to extract the topological characteristics from the SDFs of 3D shapes (Sec.~\ref{sec:top-rep}).
Within the learned latent space, we train a diffusion model with the topological features as the conditions (Sec.~\ref{sec:ddpm}). The overview of our framework is illustrated in Fig.~\ref{fig:pipeline}.

\begin{figure}
	\centering
	\includegraphics[width=1.0\linewidth]{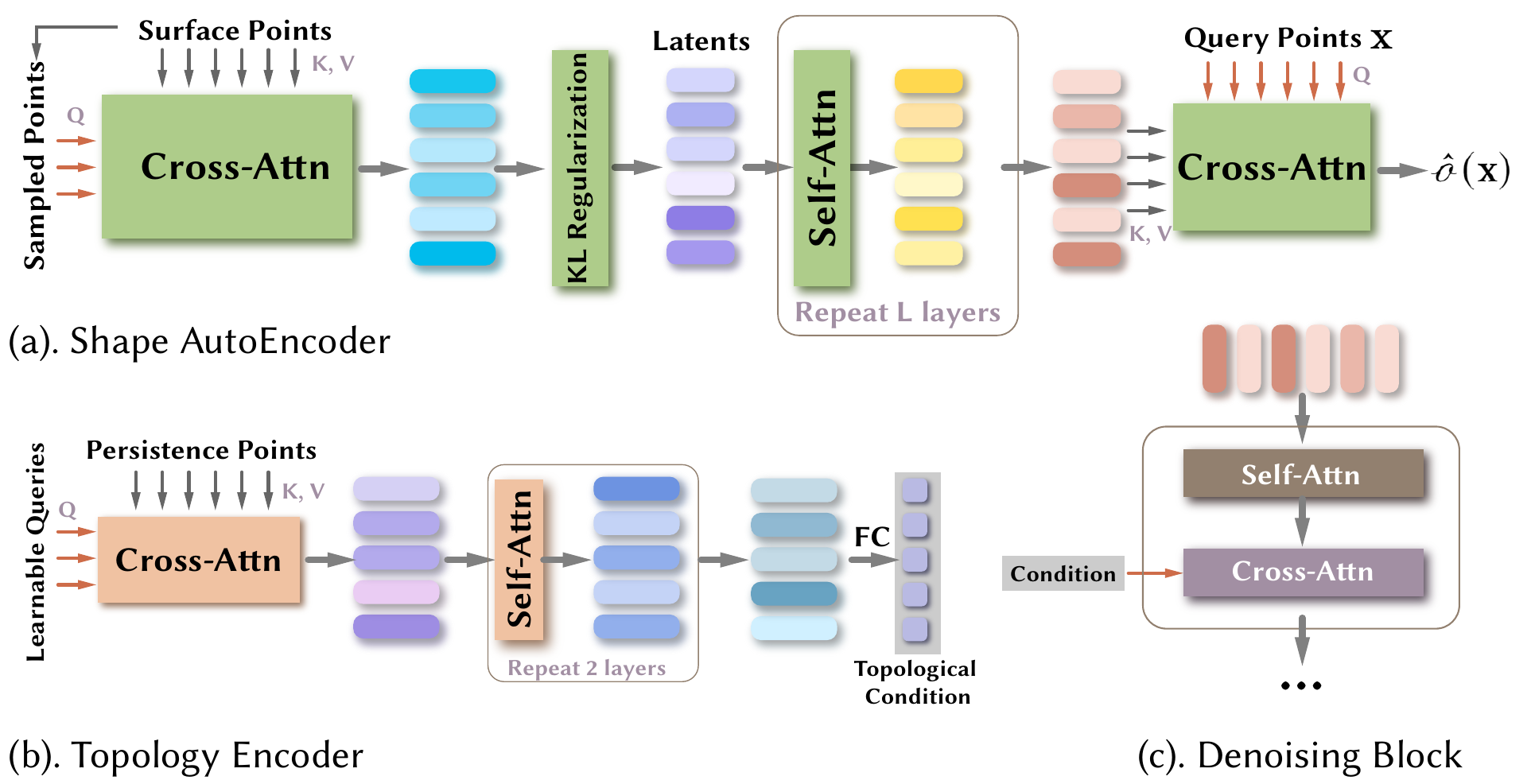}
	\caption{The architecture of Shape AutoEncoder (a), Topology Encoder (b), and Denoising Block in Diffusion process (c). We mainly utilize cross-attention (Cross-Attn) and self-attention (Self-Attn) mechanisms for the extraction and transformation of features.}
	\label{fig:subnet-arch}
\end{figure}

\subsection{Representations for 3D shapes}
\label{sec:3d-rep}
We utilize neural fields to represent 3D shapes, which is a type of neural network that takes 3D coordinates as input and outputs scalar values such as occupancy and SDF values. Similar to 3DShape2VecSet\cite{zhang20233dshape2vecset}, we encode each shape as a set of latent vectors. As shown in Fig.~\ref{fig:pipeline}(b) and Fig.~\ref{fig:subnet-arch}(a), given a dense point cloud $\mathbf{P}=\{\mathbf{p}_i\in\mathbb{R}^3\}_{i=1}^N$ sampled on the surface and a set of query points within the domain, we employ a transformer-based architecture for encoding and decoding.
\paragraph{Shape encoding.}
We choose to use the cross-attention module~\cite{jaegle2021perceiver} to encode the input point cloud $\mathbf{P}$. Specifically, we first downsample $\mathbf{P}$ to a smaller one $\tilde{\mathbf{P}}=\{\tilde{p}_i\}_{i=1}^M$, where $M\ll N$. Taking $\tilde{\mathbf{P}}$ as the query set, we apply the cross-attention mechanism as
\begin{equation}
	\mathbf{F}=\mathcal{F}_{\text{cross}}(\text{PE}(\tilde{\mathbf{P}}), \text{PE}(\mathbf{P})),
\end{equation}
where $\mathbf{F}=\{\mathbf{f}_i\in\mathbb{R}^{C}\}_{i=1}^M$ is the set of learned latent vectors, $\text{PE}:\mathbb{R}^3\rightarrow\mathbb{R}^C$ is a position embedding layer, and $\mathcal{F}_{\text{cross}}$ denotes our cross-attention module. Similarly to the approach used in latent diffusion~\cite{rombach2022high}, we also need to incorporate KL-divergence to regularize latent vectors during the generation stage. Using two linear layers, we project the latent vectors $\mathbf{f}_i$ to mean vectors $\mu_i\in\mathbb{R}^{C_0}$ and variance vectors $\log\sigma^2_i\in\mathbb{R}^{C_0}$, respectively. The VAE bottleneck can be written as $\mathbf{z}_i=\mu_i+\sigma_i\cdot\epsilon$, where $\epsilon\sim \mathcal{N}(0, 1)$. $C_0$ is much smaller than C, allowing the second stage to be trained in a smaller latent space that is easier to train. This can result in more stable 3D shapes. We write the KL regularization for the latent vectors as
\begin{equation}
	\mathcal{L}_{\text{KL}}=\frac{1}{M\cdot C_0}\sum_{i=1}^M\sum_{j=1}^{C_0}\frac{1}{2}(\mu_{i,j}^2+\sigma_{i,j}^2-\log\sigma_{i,j}^2).
\end{equation}
In our experiments, we set $C_0=32$.

\paragraph{Shape decoding.} During the decoding process, we utilize a series of self-attention modules to further process latent vectors, followed by a cross-attention module between query points and features to predict the corresponding occupancy values. Initially, we apply a linear layer $ l: \mathbb { R } ^ { C_0 } \rightarrow \mathbb { R } ^C $ to project the bottleneck $ z_i$ back into the original latent space of dimensions, obtaining $\hat{\mathbf{f}}_i\in\mathbb{R}^C$. The latent vectors are then processed by a series of self-attention layers as
\begin{equation}
	\hat{\mathbf{f}}_i^{l+1}=\mathcal{F}_{\text{self}}(\hat{\mathbf{f}}_i^{l}),\quad l=0,1,\cdots,L-1;\,\,i=1, 2, \cdots, M,
\end{equation}
where $\hat{\mathbf{f}}_i^{(0)}=\mathbf{f}_i$, $L$ is the total number of self-attention blocks, and $\mathcal{F}_{\text{self}}$ is self-attention module. Given a query point $\mathbf{x}\in\mathbf{Q}$, we can interpolate the corresponding latent vector as
\begin{equation}
	\mathbf{f}_\mathbf{x}=\sum_{i=1}^M\mathbf{v}(\hat{\mathbf{f}}^{L}_i)\bullet\frac{1}{Z_{\hat{\mathbf{x}}}}\exp(\mathbf{q}(\hat{\mathbf{x}})^\top\mathbf{k}(\hat{\mathbf{f}}^{L})/\sqrt{C}),
\end{equation}
where $\hat{\mathbf{x}}=\text{PE}(\mathbf{x})\in\mathbb{R}^C$, $Z_{\hat{\mathbf{x}}}=\sum_{i=1}^M\exp(\mathbf{q}(\hat{\mathbf{x}})^\top\mathbf{k}(\hat{\mathbf{f}}_i^L)/\sqrt{C})$ is a normalizing factor, $\mathbf{q}, \mathbf{k}$ and $\mathbf{v}$ represent the queries, keys, and values in an attention layer, respectively. Followed by a single fully connected layer, we obtain the predicted occupancy values as
\begin{equation}
	\hat{\mathbf{o}}(\mathbf{x})=\mathcal{F}_{\text{FC}}(\mathbf{x}).
\end{equation}
We employ binary cross-entropy (BCE) loss to optimize the shape AutoEncoder network as
\begin{equation}
	\mathcal{L}_{\text{BCE}}=- \frac{1}{N} \sum_{i=1}^{N} \left[ \mathbf{o}_i \cdot \log(\hat{\mathbf{o}}_i) + (1 - \mathbf{o}_i) \cdot \log(1 - \hat{\mathbf{o}}_i) \right],
\end{equation}
where $N$ is the total number of query points $\mathbf{Q}$.

\subsection{Representations for topological features}
\label{sec:top-rep}
In this work, we harness two types of topological features — Betti numbers and PDs — to guide the generation of 3D shapes. Our focus is primarily on the loop features of shapes, specifically the 1-dimensional Betti number $ \beta _1 $ and the 1-dimensional PD $ \mathcal { D } ^1 $. It is worth noting that our proposed framework can be extended to include consideration of the number of components and voids, provided that a sufficient data set is available.

\paragraph{Betti number.}
As a topological invariant, the Betti number can be regarded as a discrete feature to classify shapes. To make it adaptable to the network, we map the Betti number $\beta^1$ to a continuous vector $\mathbf{c}_{\beta}$ by an embedding layer.

\begin{figure}[htbp]
	\centering
	\includegraphics[width=0.9\linewidth]{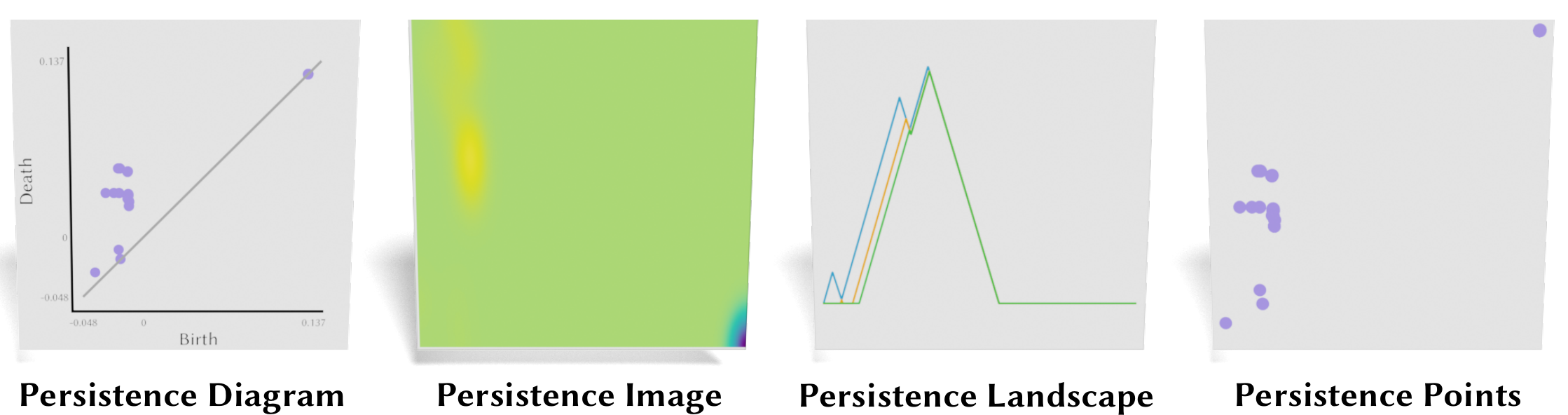}
	\caption{Distinct from the vectorization approaches of persistence images and persistence landscapes, our method interprets the birth-death pairs from persistence homology (PH) analysis as discrete points within a 2D space. This representation serves as the underpinning for our topological feature characterization. }
	\label{fig:topo-rep}
\end{figure}

\paragraph{Persistence points.}
In order to efficiently use persistent homology features within neural networks, many methods replace PD with network-friendly finite-dimensional vector representations, such as Persistence Landscape (PL)~\cite{bubenik2015statistical} and Persistence Image (PI)~\cite{adams2017persistence}, as shown in Fig.~\ref{fig:topo-rep}. However, there are several issues with using this method as conditional information to guide the training of generative models. The process of vectorization necessitates the definition of multiple parameters, which poses a challenge in attaining a consistent vector representation for extensive data. Furthermore, the vectorization process is irreversible, and subsequent utilization of networks like CNN for vector analysis will also result in additional loss of information. Therefore, we use the persistence points $g_i=(b_i^1, d_i^1-b_i^1)\in\mathcal{D}^1$ directly as features and employ a Transformer-based encoder $\mathcal{T}$ (as shown in Fig.~\ref{fig:subnet-arch}(a)) to extract topological conditions. That is,
\begin{equation}
	\mathbf{c}_{\text{PD}}=\mathcal{T}(\mathbf{G}), \quad \mathbf{G}=\{g_i\}.
\end{equation}
In our experiments, we set $\mathbf{c}_{\beta}\in\mathbb{R}^{256}$ and $\mathbf{c}_{\text{PD}}\in\mathbb{R}^{256}$.

\subsection{Topology-aware shape generation}
\label{sec:ddpm}

After acquiring the compressed latent representation (i.e., bottleneck $ z_i $ ) of 3D shapes, we can generate new shapes by diffusing in the latent space. EDM~\cite{karras2022elucidating} understands diffusion models from the perspective of score matching~\cite{hyvarinen2005estimation} and proposes a novel framework to train the diffusion process. At the noise level of $\sigma$, we add noise to $z_i$ as $z_i+\mathbf{n}_i$, where $\mathbf{n}_i\sim\mathcal{N}(\mathbf{0}, \sigma^2\mathbf{I})$. We use a network to learn denoising under the following loss,
\begin{equation}
	\mathcal{L}_{\text{EDM}}=\mathbb{E}_{\mathbf{n}_i}\frac{1}{M}\sum_{i=1}^M\left\|\mathcal{F}_{D}(z_i+\mathbf{n}_i, \sigma, \mathbf{c})-z_i\right\|,
\end{equation}
where $\mathcal{F}_D$ is our denoising network, $\mathbf{c}$ is the conditional vectors. As shown in Fig.~\ref{fig:subnet-arch}(b), each denoising block is composed of one self-attention layer and one cross-attention layer. We can incorporate the topological conditions $ \mathbf { c } _ \beta $ or $ \mathbf { c } _ \text { PD } $ to achieve the topology-aware generation. During the sampling process, we can explore the learned latent space of topological conditions to create a wide variety of 3D shapes. This can involve specifying Betti numbers or randomly sampling $ \mathbf { c } _ { \text { PD } } $.

\section{Experimental Setup}
\label{sec:setup}

\subsection{Baselines \& evaluation metrics}
For shape generation, we compare against recent state-of-the-art 3D generative models, including SDF-StyleGAN~\cite{zheng2022sdf}, NFD~\cite{shue20233d}, GET3D~\cite{gao2022get3d}, MeshDiffusion~\cite{liu2023meshdiffusion}, SLIDE~\cite{lyu2023controllable}, 3DILG~\cite{zhang20223dilg}, and LAS-Diffusion~\cite{zheng2023locally}.
Compared to other 3D shape generation models, our approach enhances the diversity of the generated shapes by introducing topological features. To demonstrate this, we employ the Shading-FID and 1-NNA metrics to quantitatively compare the performance of these methods.
Following the procedures outlined in~\cite{zheng2022sdf}, we rendered multi-view images of the meshes in the generated set $S_g$ reference set $S_r$ (including the data for training, validation, and testing). Then we compute the Fréchet inception distance (FID) score based on their $i$-th view images, and average 20 FID scores to define the FID as:
\begin{equation}
	F I D=\frac{1}{20}\left[\sum_{i=1}^{20}\left\|\mu_g^i-\mu_r^i\right\|^2+\operatorname{Tr}\left(\Sigma_g^i+\Sigma_r^i-2\left(\Sigma_r^i \Sigma_g^i\right)^{1 / 2}\right)\right],
\end{equation}
where $g$ and $r$ denote the features of the generated data set and the training set, $\mu^i, \Sigma^i$. denote the mean and the covariance matrix of the corresponding shading images rendered from the $i$-th view.

We also compare the 1-nearest neighbor accuracy (1-NNA) metric to evaluate the diversity, which directly quantifies distribution similarity between the reference set $S_r$ and the generated set $S_g$.
\begin{equation}
	1-\mathrm{NNA}\left(S_g, S_r\right)=\frac{\sum_{X \in S_g} \mathbb{I}\left[N_X \in S_g\right]+\sum_{Y \in S_r} \mathbb{I}\left[N_Y \in S_r\right]}{\left|S_g\right|+\left|S_r\right|},
	\label{1nna}
\end{equation}
where $\mathbb{I}[\cdot]$ is the indicator function and $N_X$ is the nearest neighbor of $X$ in the set $S_r \cup$ $S_g-\{X\}$. We evaluate the 1-NNA based on both Chamfer Distance (CD) and Earth Mover's Distance (EMD).
For 1-NNA, 50 $\%$ indicates the distribution of generated meshes is close to the reference set. A higher value of 1-NNA signifies greater diversity.
Furthermore, we employ CD/EMD-based Coverage (COV) as a metric to quantify diversity, with a higher value indicating a greater level of diversity.

\subsection{Implementation}
We select 6778 meshes of the chair category and 8436 meshes of the table category from the ShapeNet data set~\cite{shapenet2015}, and select 7731 CAD meshes from the ABC dataset~\cite{Koch_2019_CVPR}. We normalize all the meshes to a bounding box with the range of $[-0.4, 0.4]$. Each category of the dataset is split into training, validation and test sets with a ratio of 7:1:2, respectively. To obtain implicit field representations of 3D shapes, we voxelized the occupied bounding box with the range $[-0.5, 0.5]$ with a resolution of $128^3$. We trained our models on 6 A100 with batch size of 32.
\paragraph{Persistent homology analysis} We calculated the SDF values at the voxel points~\cite{wang2022dual}. Subsequently, utilizing the cubical complex and taking the SDF as filtration values, we conducted the PH analysis to obtain the PDs with the GUDHI library~\cite{maria2014gudhi}.
Considering that most points near the diagonal line are noises, we only selected the 16 points with the longest persistence to input into the Toplogy Encoder $\mathcal{T}$ to extract topological conditions.
\paragraph{Shape autoencoding}
For the shape AutoEncoder, we sample 2048 points $\mathbf{P}$ on the surface to input the encoder and individually sample 2048 query points from the bounding box ($[-0.5, 0.5]^3$) at each iteration. We trained the shape AutoEncoder for 1000 epochs. Similar to 3DShape2VecSet~\cite{zhang20233dshape2vecset}, we configured the learning rate to linearly increase to $5e-5$ in the first 80 epochs, and then it was gradually decreased to $1e-6$ following the cosine decay schedule.
\paragraph{Diffusion model}
The diffusion model consists of 24 denoising blocks. We conducted the training the diffusion process for 1500 epochs. The learning rate is linearly increased to $1e-4$ in the first 100 epochs, and then decreased to $1e-6$ following the cosine decay schedule. For the configuration of additional hyperparameters, please refer to EDM~\cite{karras2022elucidating}. To incorporate sparse point clouds as conditions, we sample 256 points from the dense point clouds $\mathbf{P}$. When working with partial point clouds, we randomly crop a quarter of $\mathbf{P}$. We use the PointNet~\cite{qi2017pointnet} as an encoder to embed these point clouds into a condition vector $\hat{\mathbf{c}} \in \mathbb{R}^{256}$. For the sketch images, we utilize the ViT backbone~\cite{dosovitskiy2020image} as the encoder. After obtaining the condition vector $\hat{\mathbf{c}}$ from point clouds or sketches, we concatenate them with the topological conditions and input them together to each denoising block.
\begin{figure}[ht]
	\centering
	\makebox[0.02\linewidth]{\rotatebox{90}{$\beta_1=0$}}
	\includegraphics[width=0.15\linewidth]{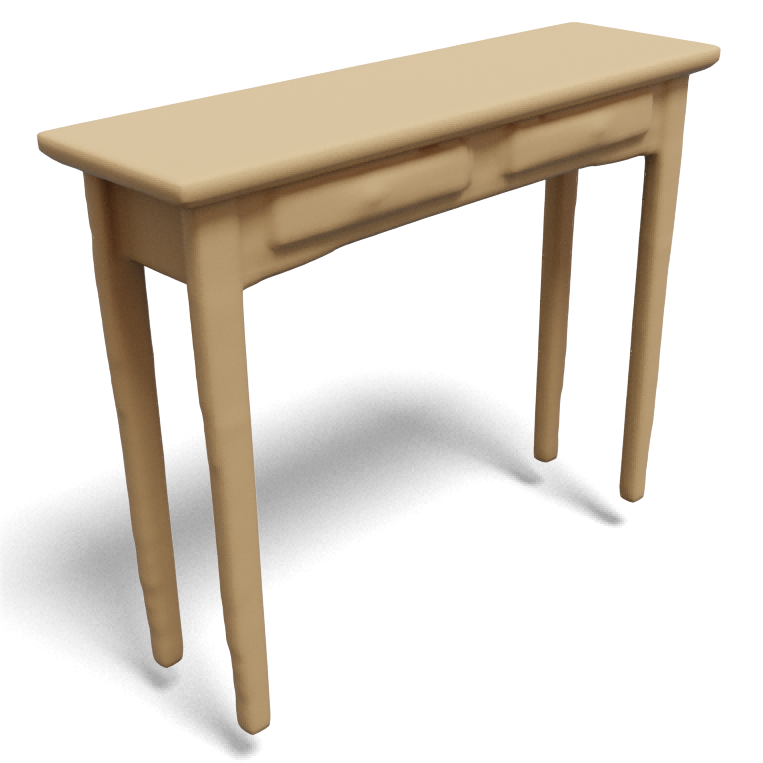}
	\includegraphics[width=0.15\linewidth]{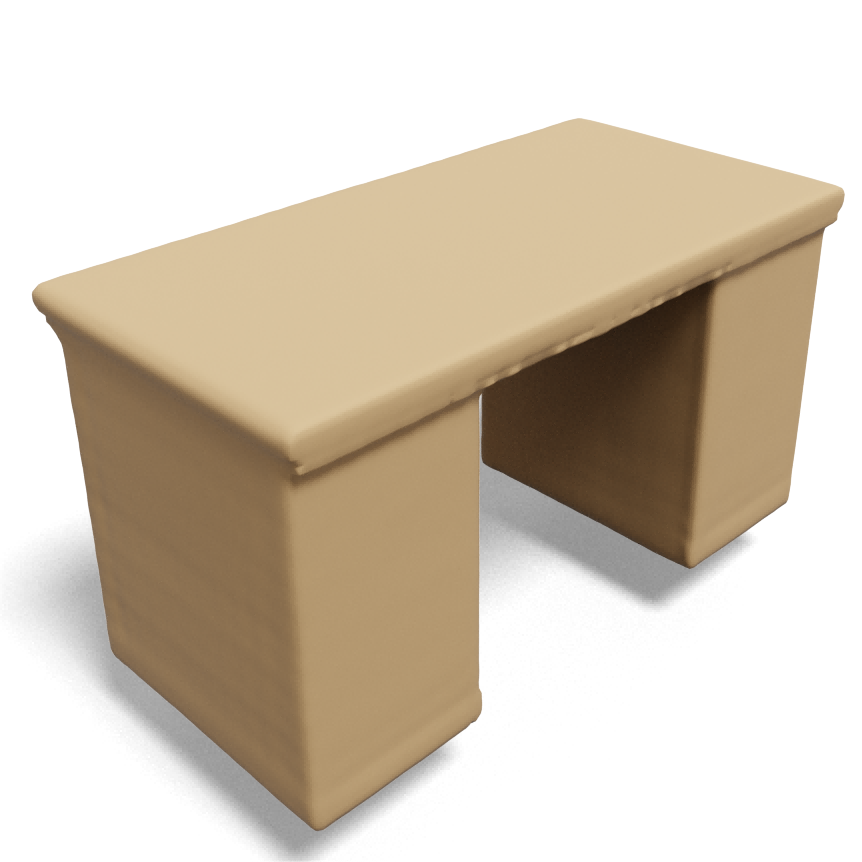}
	\includegraphics[width=0.15\linewidth]{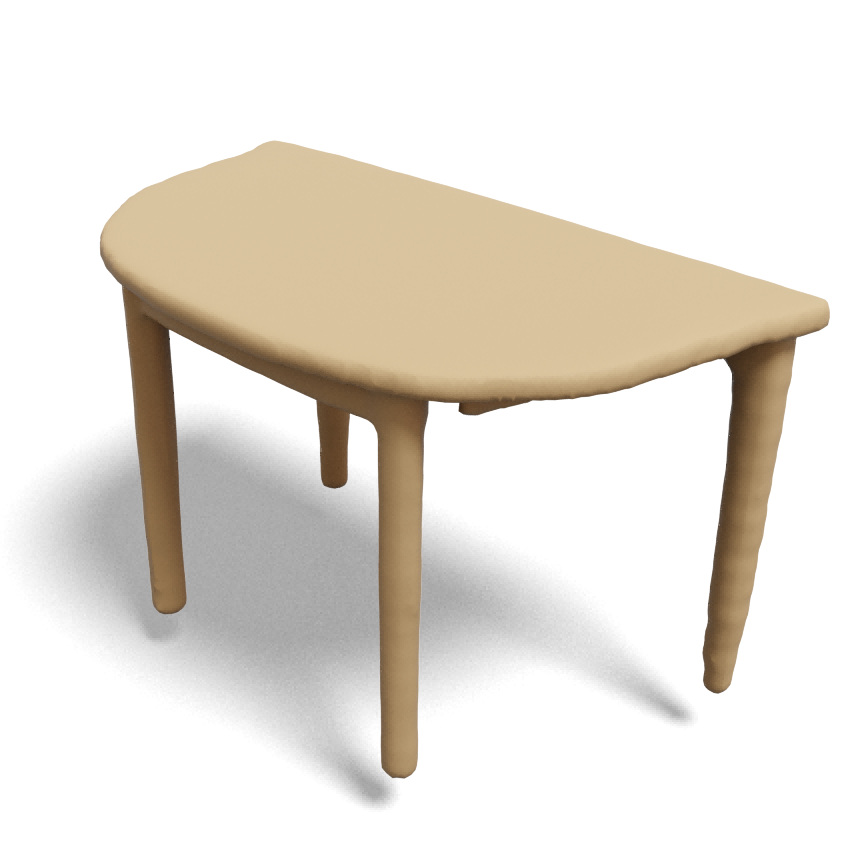}
	\includegraphics[width=0.15\linewidth]{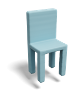}
	\includegraphics[width=0.15\linewidth]{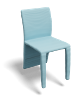}
	\includegraphics[width=0.15\linewidth]{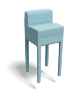}
	\vspace{-5pt}
	\\
	\makebox[0.02\linewidth]{\rotatebox{90}{$\beta_1=1$}}
	\includegraphics[width=0.15\linewidth]{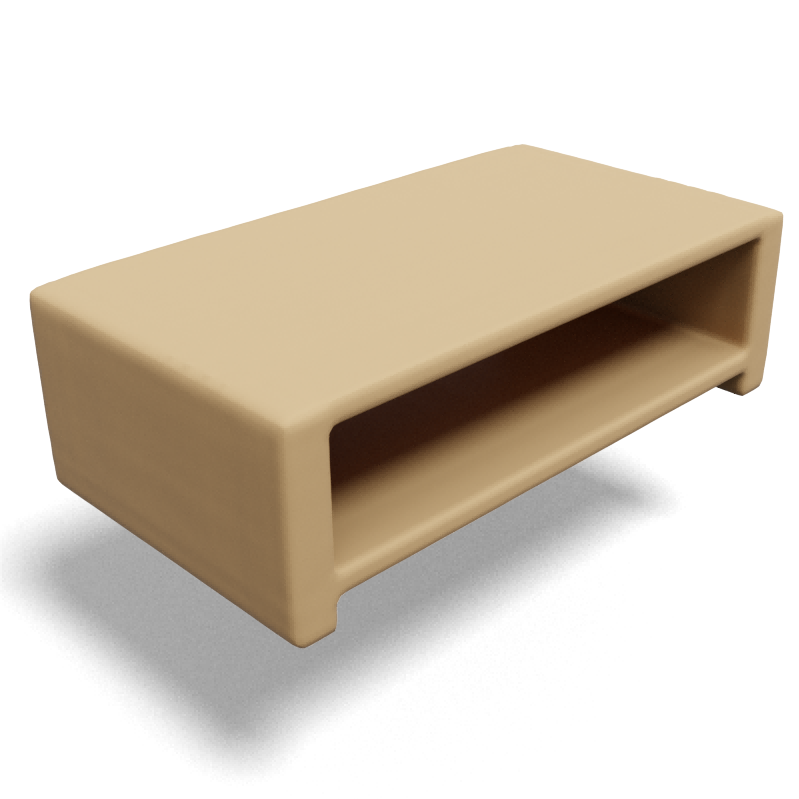}
	\includegraphics[width=0.15\linewidth]{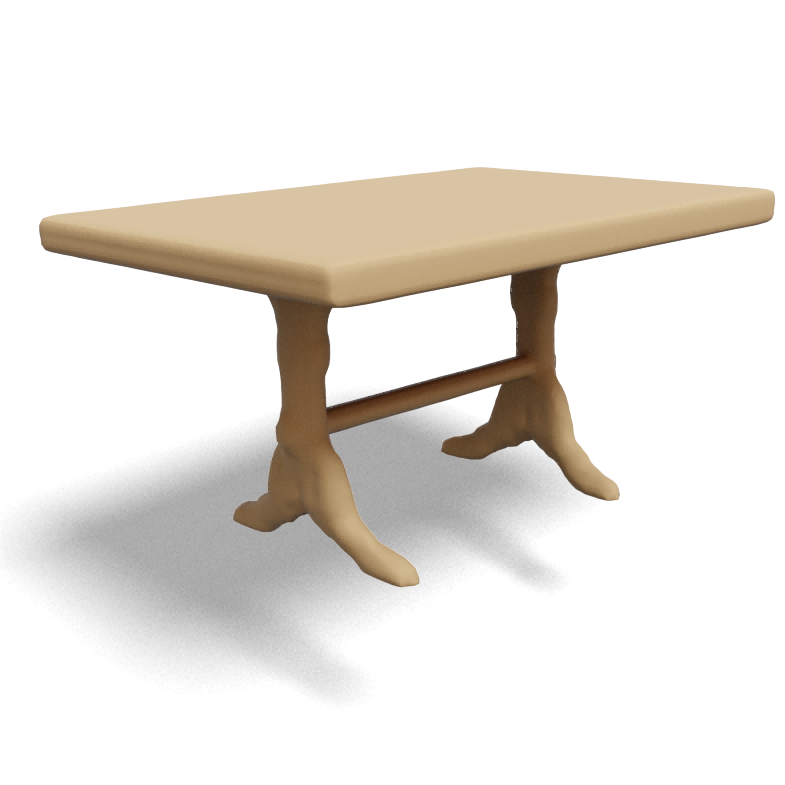}
	\includegraphics[width=0.15\linewidth]{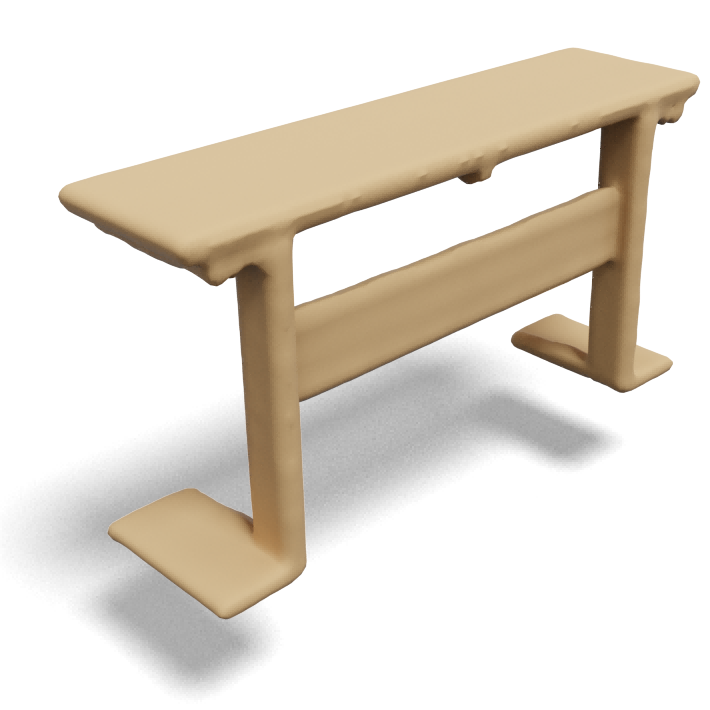}
	\includegraphics[width=0.15\linewidth]{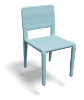}
	\includegraphics[width=0.15\linewidth]{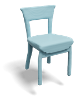}
	\includegraphics[width=0.15\linewidth]{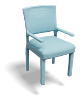}
	\vspace{-5pt}
	\\
	\makebox[0.02\linewidth]{\rotatebox{90}{$\beta_1=2$}}
	\includegraphics[width=0.15\linewidth]{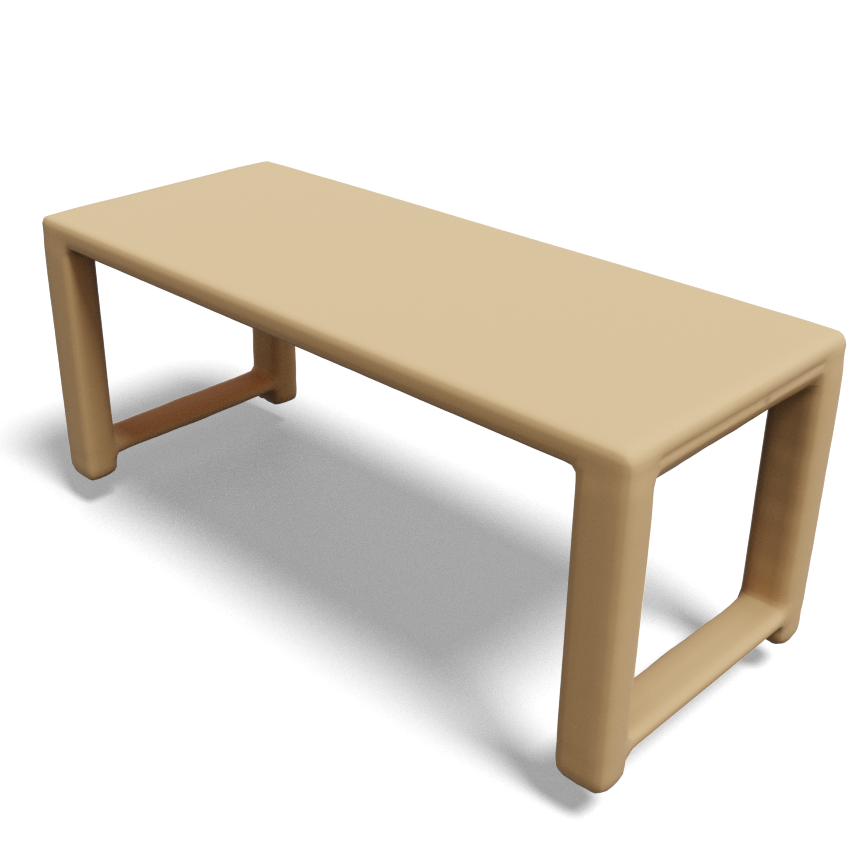}
	\includegraphics[width=0.15\linewidth]{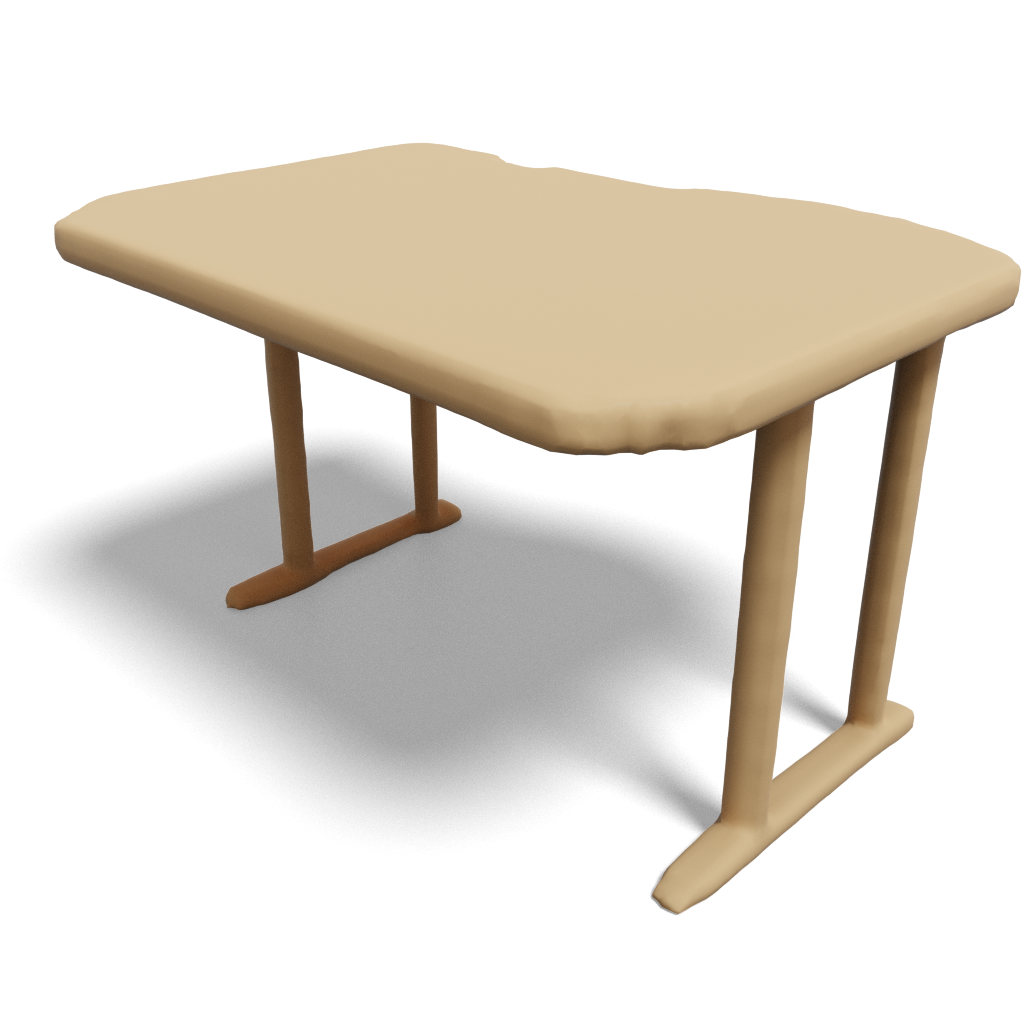}
	\includegraphics[width=0.15\linewidth]{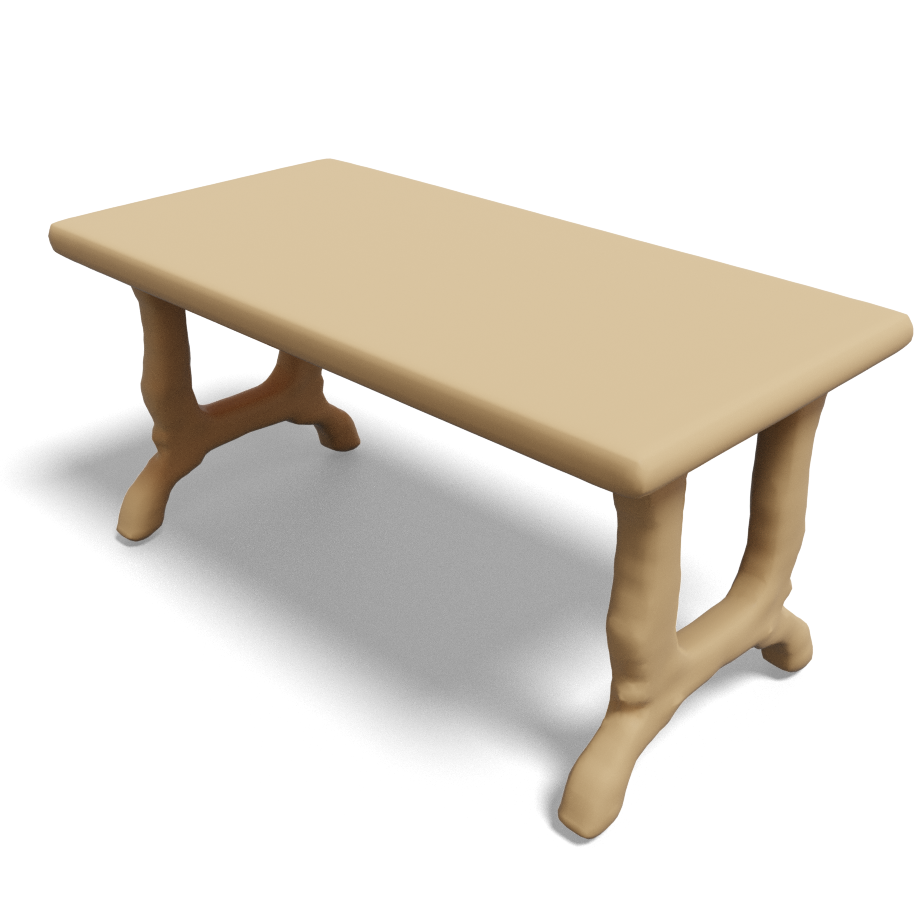}
	\includegraphics[width=0.15\linewidth]{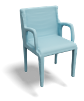}
	\includegraphics[width=0.15\linewidth]{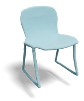}
	\includegraphics[width=0.15\linewidth]{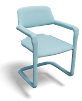}
	\vspace{-5pt}
	\\
	\makebox[0.02\linewidth]{\rotatebox{90}{$\beta_1=3$}}
	\includegraphics[width=0.15\linewidth]{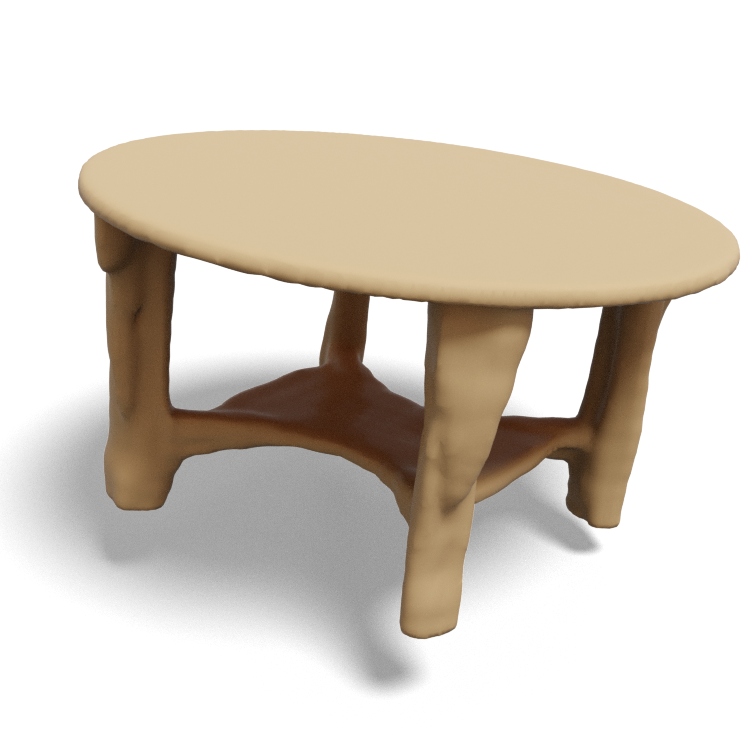}
	\includegraphics[width=0.15\linewidth]{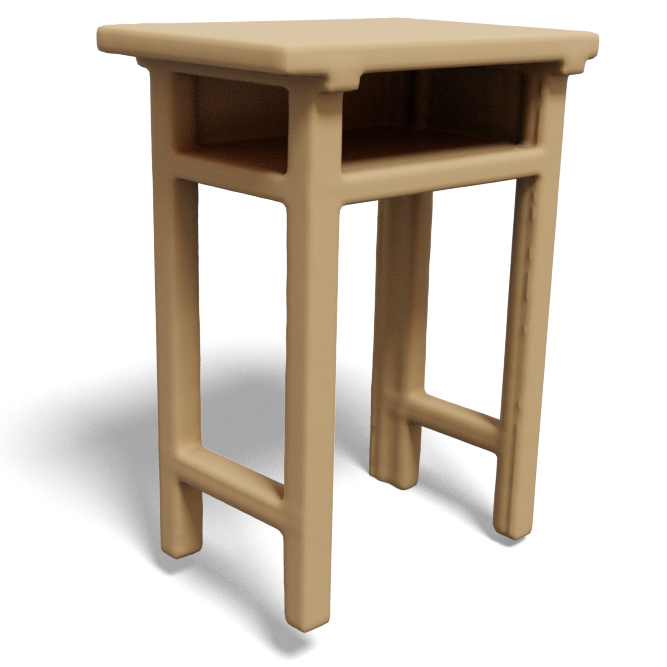}
	\includegraphics[width=0.15\linewidth]{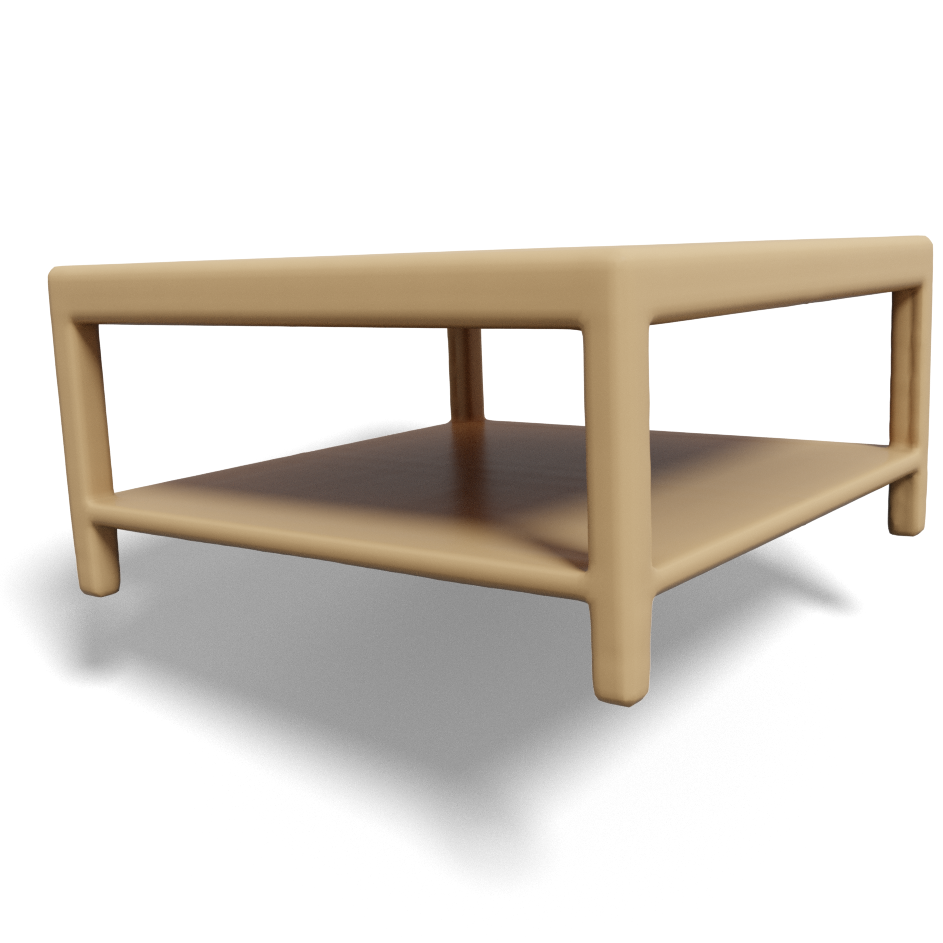}
	\includegraphics[width=0.15\linewidth]{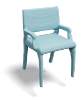}
	\includegraphics[width=0.15\linewidth]{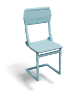}
	\includegraphics[width=0.15\linewidth]{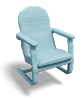}
	\vspace{-5pt}
	\\
	\makebox[0.02\linewidth]{\rotatebox{90}{$\beta_1=4$}}
	\includegraphics[width=0.15\linewidth]{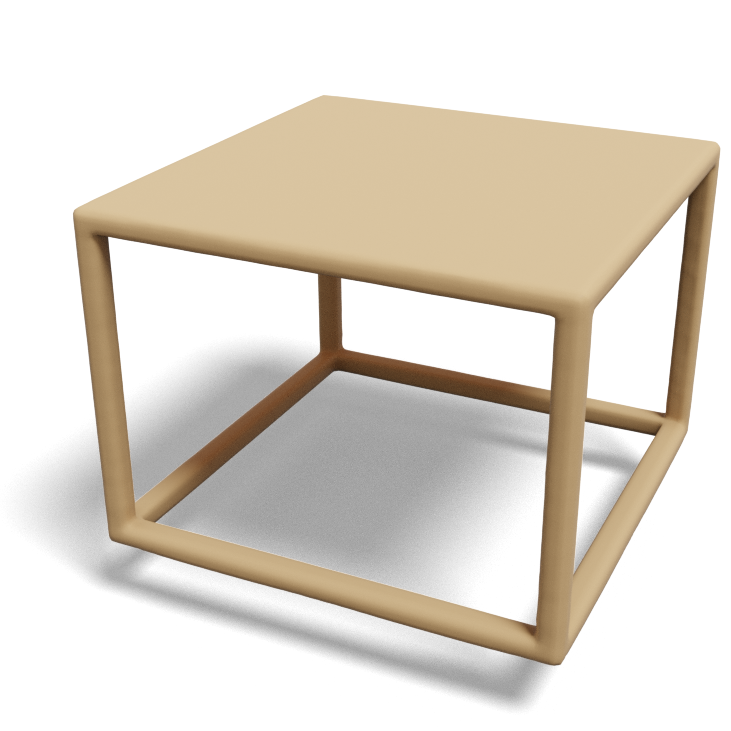}
	\includegraphics[width=0.15\linewidth]{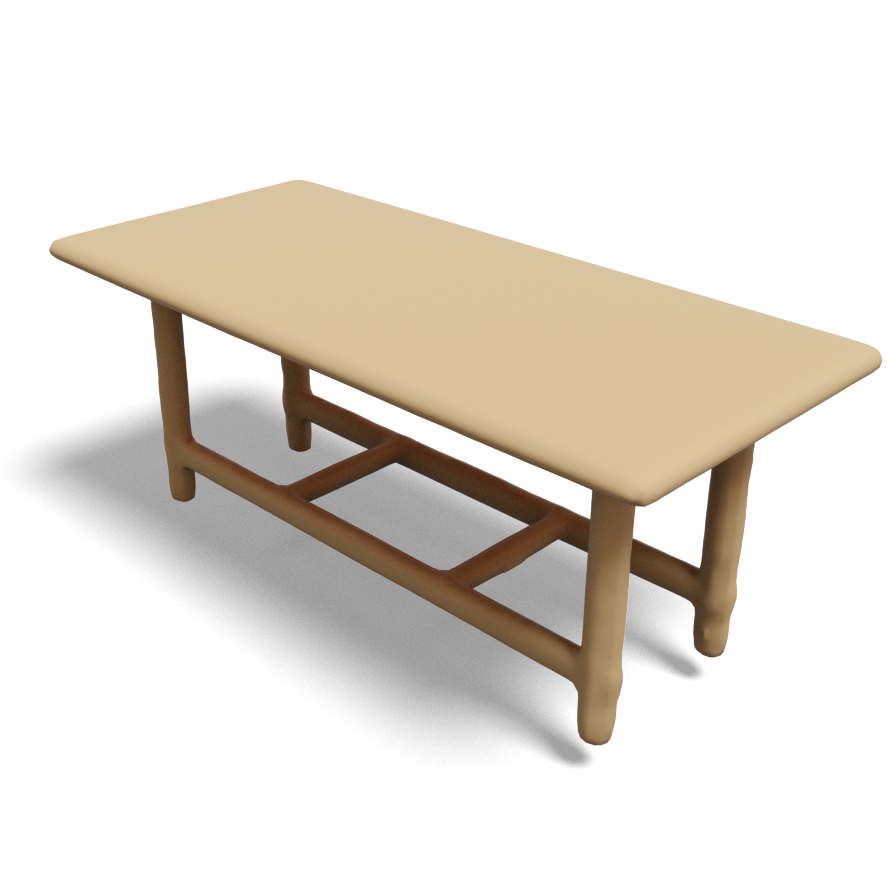}
	\includegraphics[width=0.15\linewidth]{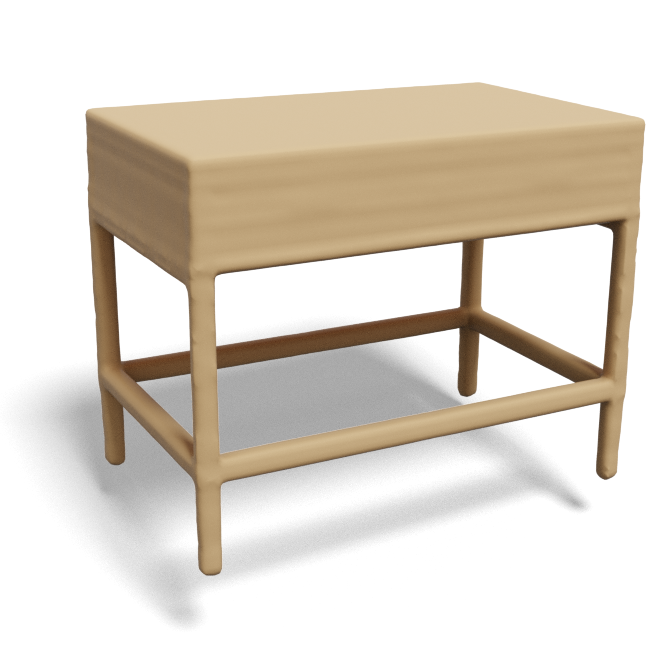}
	\includegraphics[width=0.15\linewidth]{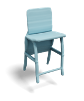}
	\includegraphics[width=0.15\linewidth]{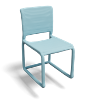}
	\includegraphics[width=0.15\linewidth]{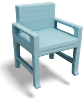}
	\caption{We can generate 3D shapes with different topologies by specifying the 1-dimension Betti number $\beta_1$.}
	\label{fig:betti-cond}
\end{figure}

\begin{figure*}[hbp]
	\centering
	\includegraphics[width=1.0\linewidth]{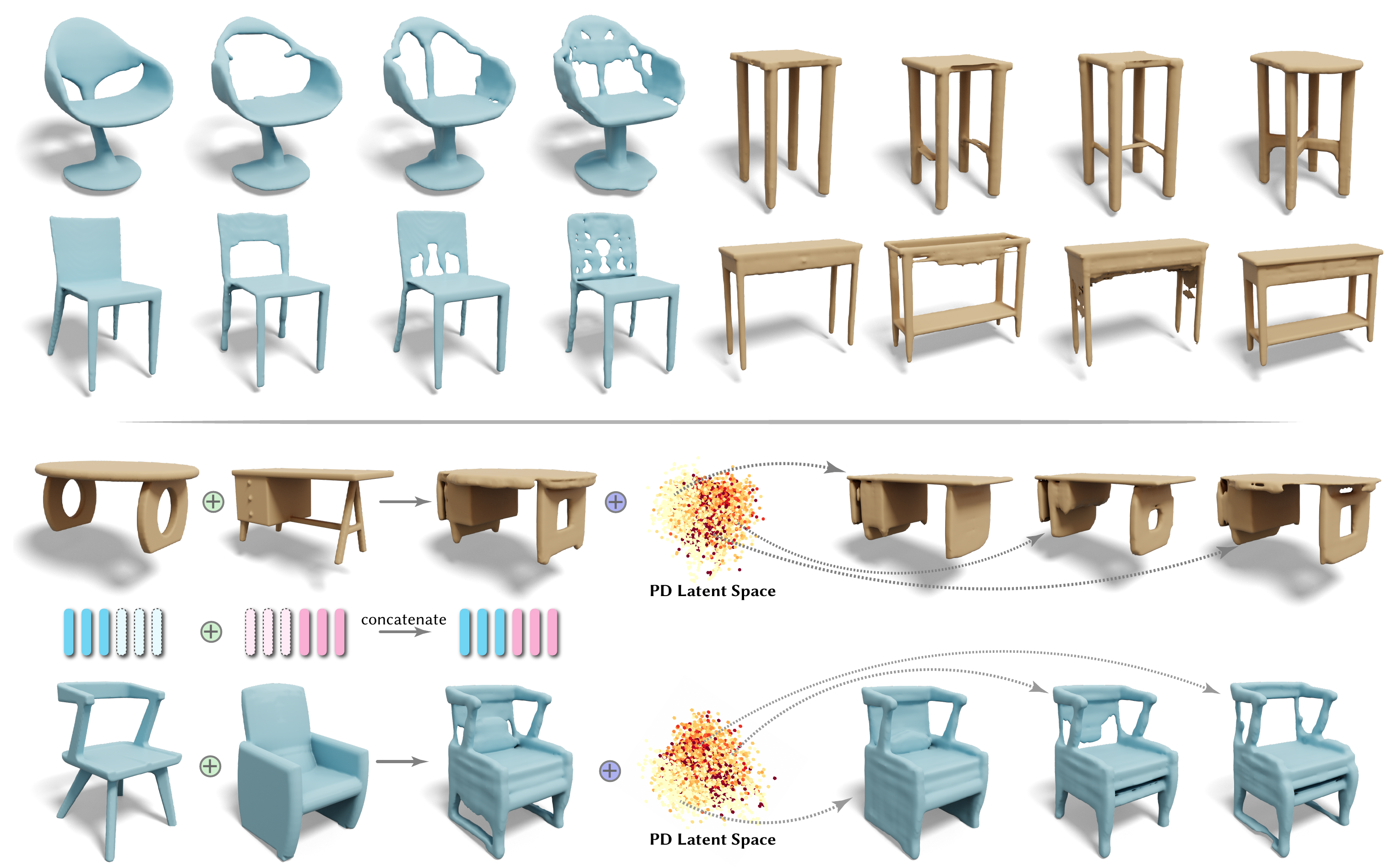}
	\caption{By sampling more topological features from the learned PD latent space, we can increase the diversity of the 3D shapes generated (Top). Another method to obtain new shapes is by merging the latent sets of two shapes, and then further enriching the shapes by sampling from the PD latent space (Bottom). }
	\label{fig:pd-cond}
\end{figure*}

\begin{figure*}[htbp]
	\centering
	\includegraphics[width=1.0\linewidth]{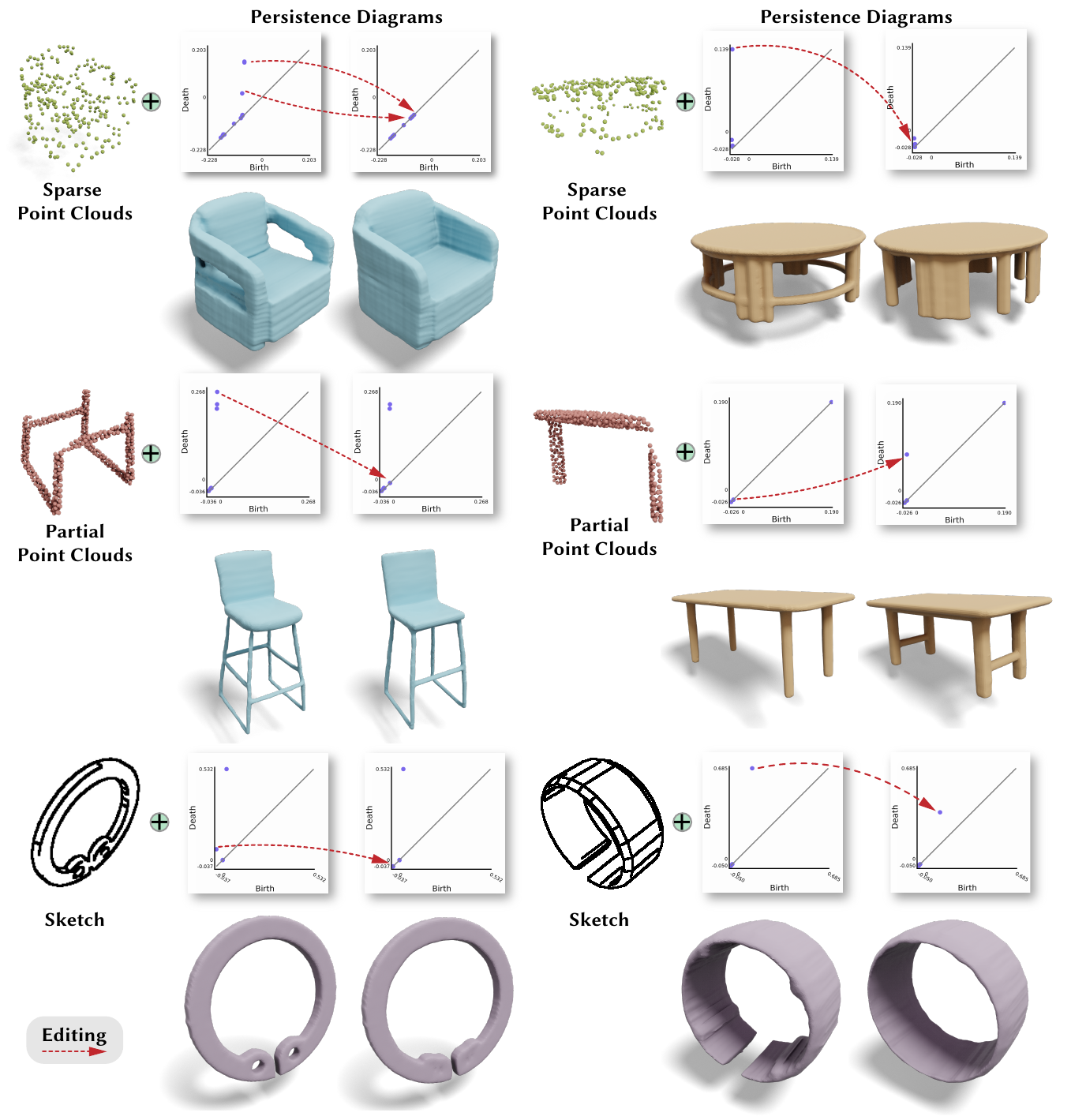}
	\caption{Our approach enables the shape generation from different types of inputs, such as sparse/partial point clouds and sketches. Furthermore, we can alter the topology of the generated shapes by modifying the PDs.}
	\label{fig:pd-ctrl}
\end{figure*}

\section{Results}
\label{sec:results}

Since the diffusion training is under the guidance of topological features, we can explore the learned topological space during the sampling process to enhance the diversity of generated shapes.
As shown in Fig.~\ref{fig:betti-cond}, we can generate 3D shapes with different topologies by specifying the 1-dimensional Betti numbers $\beta_1$. The final generated shapes possess varying loop features. However, serving as discrete topology invariants, Betti numbers have relatively small degrees of freedom (owing to limitations of the available data set, we only considered cases where $ \beta_1 \leq 4 $ in our experiments). In contrast, the persistence points induced from the SDFs are densely distributed in the feature space. We can consider it as a continuous space from which sampling yields a wide variety of topological features.
As shown in Fig.~\ref{fig:pd-cond}, we can generate more complex shapes by sampling the topological features from the PD latent space learned by the Topology Encoder.
We compare the FID and 1-NNA (CD/EMD) metrics with other generative methods in Tab.~\ref{table:comparison}.
Our method without topological features achieves comparable FID and 1-NNA with other generative methods, demonstrating that relatively high-quality meshes are generated.
When integrating topological features, our method fulfills the highest FID in the Chair and Table, as well as the second highest 1-NNA (CD/EMD), proving that topological features enable our method with more diversity.

\begin{table*}[ht]
    \resizebox{\textwidth}{!}
    {
        \begin{tabular}{l|ccc|ccc}
            \toprule[1.5pt]
            \multirow{2}{*}{} & \multicolumn{3}{c|}{Chair} & \multicolumn{3}{c}{Table}                                                                                               \\ \cline{2-7}
                              & FID$\uparrow$              & 1-NNA $\uparrow$ (CD/EMD) & COV$\uparrow$ (CD/EMD) & FID$\uparrow$  & 1-NNA$\uparrow$ (CD/EMD) & COV$\uparrow$ (CD/EMD) \\ \midrule[1pt]
            SDF-StyleGAN      & 46.04                      & 65.22/70.27               & 43.24/38.44            & 45.57          & 75.18/75.08              & 38.14/35.84            \\
            NFD               & 40.39                      & 56.36/57.20               & 45.35/44.94            & -              & -                        & -                      \\
            GET3D             & 66.48                      & 68.97/65.16               & 45.95/47.75            & 64.06          & 68.37/67.76              & 44.04/45.95            \\
            MeshDiffusion     & 76.81                      & 67.77/67.86               & 40.14/42.54            & 79.59          & 57.81/66.76              & 49.05/45.25            \\
            SLIDE             & 43.64                      & 59.41/61.01               & 46.75/46.65            & -              & -                        & -                      \\
            3DILG             & 29.71                      & \textbf{74.27/73.62}      & 36.54/37.84            & 52.97          & \textbf{82.43/81.43}     & 24.22/27.23            \\ \midrule[1pt]
            Ours (w/o Topo)   & 63.84                      & 58.84/58.89               & 49.21/46.98            & 42.73          & 59.15/59.20              & 49.21/50.37            \\
            Ours (w/ Topo)    & \textbf{96.54}             & 69.77/67.07               & \textbf{49.54/51.07}   & \textbf{92.53} & 79.22/77.41              & \textbf{49.54/51.07}   \\ \bottomrule[1.5pt]
        \end{tabular}
    }
    \caption{Shading FID scores, 1-NNA, and COV comparison between our method and baselines. ``-'' indicates that no checkpoint is provided.}
    \label{table:comparison}
\end{table*}

\begin{figure}[ht]
	\centering
	\includegraphics[width=1.0\linewidth]{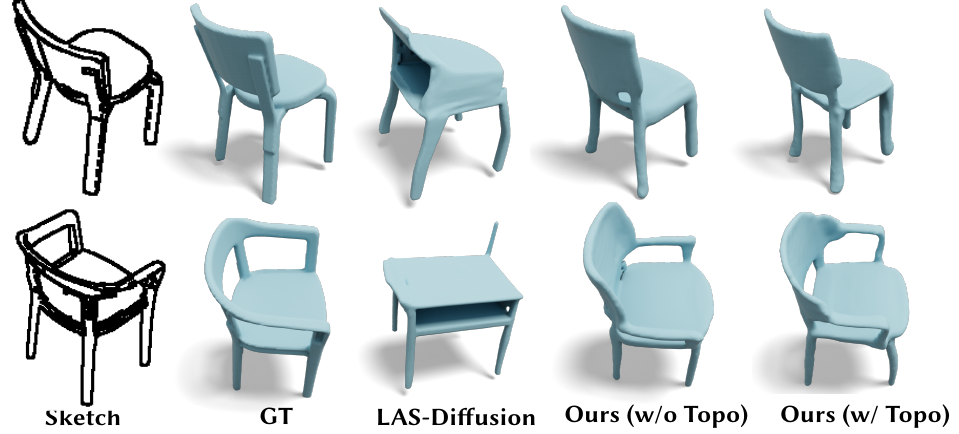}
	\caption{Comparison with sketch-based generative method. LAS-Diffusion leverages local attention while overlooking global topological constraints, leading to incorrect results on unseen parts. After considering global topological characteristics, our method can generate high-quality meshes that are faithful to the sketch condition.}
	\label{fig:compare-las}
\end{figure}

Furthermore, our framework enables shape generation from different types of input, including sparse and partial point clouds, as well as sketches. Nevertheless, the inherent sparseness and lack of visibility in these inputs can lead to topological uncertainties, which pose challenges in generating shapes with the desired structures. Our topology-aware generative model is capable of altering the topology of the generated shapes by adjusting the positions of persistence points on the PDs. The guiding rule for alterations is that as a point is drawn nearer to the diagonal line, the lifespan of the related topological characteristic shortens, leading it toward disappearance. If all points were positioned on the diagonal, then the shape would correspond to a shape without any holes, known as a genus zero object. Fig.~\ref{fig:pd-ctrl} illustrates the effectiveness of our approach under various inputs.
LAS-Diffusion~\cite{zheng2023locally} can generate controllable 3D shapes from sketches utilizing the local attention mechanism. However, this method struggles to restore the correct structural features when they are not apparent in the sketch, due to its lack of consideration for global structural characteristics. We compare the generated results in Fig.~\ref{fig:compare-las}, it can be seen that our method is capable of obtaining 3D shapes from sketches with the desired topology.

\begin{figure}[ht]
	\centering
	\includegraphics[width=1.0\linewidth]{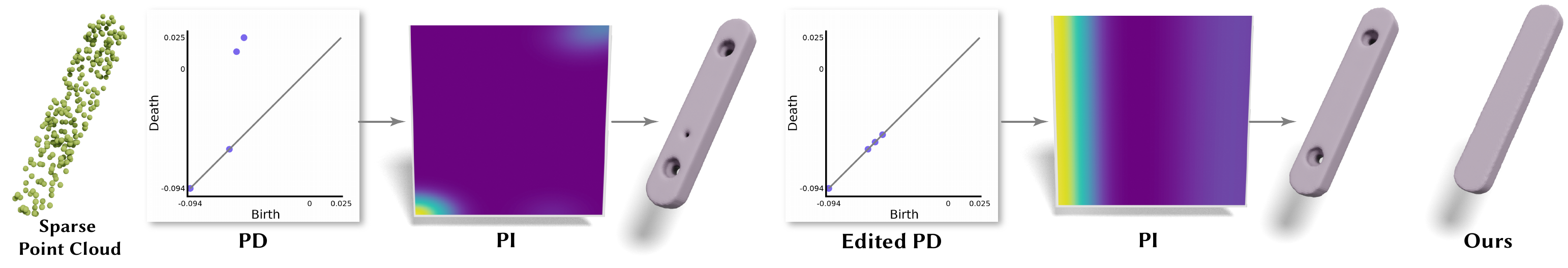}
	\caption{Although persistence image (PI) is a network-friendly representation for persistence diagram, it struggles to control the topology characteristics of the generated shapes.}
	\label{fig:abla}
\end{figure}

\paragraph{Ablation study}
As depicted in Sec.~\ref{sec:top-rep}, existing network-based methods utilize the persistence image (PI) or persistence landscape (PL) to replace PDs in their architectures.  Our research is the first to directly extract topological features from the points of PDs by using a transformer-based topological encoder. We conducted experiments that involved utilizing PI as the topological feature input and employing a Vision Transformer (ViT) backbone~\cite{dosovitskiy2020image} for feature extraction, which then served as the topological condition for training diffusion models. As shown in Fig.~\ref{fig:abla}, it is challenging to modify the topology of the generated shapes by altering the PI due to the loss of information resulting from intermediary transformations.

\section{Conclusions}
In this study, we introduce a topology-aware latent diffusion model tailored for the generation of 3D shapes. Leveraging topological features derived from persistent homology analysis, we condition the diffusion process within a latent space learned by a transformer-based AutoEncoder. Our experimental results showcase the model's effectiveness in producing a diverse array of 3D shapes, each with distinct topological properties. Additionally, our framework exhibits versatility by facilitating shape generation from varied inputs, including sparse or partial point clouds and sketches. A unique aspect of our approach is the ability to adjust the topology of the generated shapes by modifying the persistence diagrams, offering an advancement in controllable 3D shape synthesis.

\paragraph{Limitations\& Future work}
One limitation of our method is the dependence on the dataset. It requires providing a large number of datasets with different topological features for training in order to produce relatively stable results. In theory, as long as there is an available dataset, our method can be applied to the generation of 3D geometries of any non-watertight and watertight surfaces.

\clearpage
\newpage
\bibliographystyle{unsrtnat}
\bibliography{references}

\begin{thebibliography}{60}
\providecommand{\natexlab}[1]{#1}
\providecommand{\url}[1]{\texttt{#1}}
\expandafter\ifx\csname urlstyle\endcsname\relax
  \providecommand{\doi}[1]{doi: #1}\else
  \providecommand{\doi}{doi: \begingroup \urlstyle{rm}\Url}\fi

\bibitem[Wang et~al.(2021)Wang, Liu, Liu, Theobalt, Komura, and Wang]{wang2021neus}
Peng Wang, Lingjie Liu, Yuan Liu, Christian Theobalt, Taku Komura, and Wenping Wang.
\newblock Neus: Learning neural implicit surfaces by volume rendering for multi-view reconstruction.
\newblock \emph{arXiv preprint arXiv:2106.10689}, 2021.

\bibitem[Yariv et~al.(2021)Yariv, Gu, Kasten, and Lipman]{yariv2021volume}
Lior Yariv, Jiatao Gu, Yoni Kasten, and Yaron Lipman.
\newblock Volume rendering of neural implicit surfaces.
\newblock \emph{Advances in Neural Information Processing Systems}, 34:\penalty0 4805--4815, 2021.

\bibitem[Xu et~al.(2023{\natexlab{a}})Xu, Mu, and Yang]{xu2023survey}
Qun-Ce Xu, Tai-Jiang Mu, and Yong-Liang Yang.
\newblock A survey of deep learning-based 3d shape generation.
\newblock \emph{Computational Visual Media}, 9\penalty0 (3):\penalty0 407--442, 2023{\natexlab{a}}.

\bibitem[Groueix et~al.(2018)Groueix, Fisher, Kim, Russell, and Aubry]{groueix2018papier}
Thibault Groueix, Matthew Fisher, Vladimir~G Kim, Bryan~C Russell, and Mathieu Aubry.
\newblock A papier-m{\^a}ch{\'e} approach to learning 3d surface generation.
\newblock In \emph{Proceedings of the IEEE conference on computer vision and pattern recognition}, pages 216--224, 2018.

\bibitem[Zheng et~al.(2022)Zheng, Liu, Wang, and Tong]{zheng2022sdf}
Xinyang Zheng, Yang Liu, Pengshuai Wang, and Xin Tong.
\newblock Sdf-stylegan: Implicit sdf-based stylegan for 3d shape generation.
\newblock In \emph{Computer Graphics Forum}, volume~41, pages 52--63. Wiley Online Library, 2022.

\bibitem[Chen et~al.(2021)Chen, Kim, Fisher, Aigerman, Zhang, and Chaudhuri]{chen2021decor}
Zhiqin Chen, Vladimir~G Kim, Matthew Fisher, Noam Aigerman, Hao Zhang, and Siddhartha Chaudhuri.
\newblock Decor-gan: 3d shape detailization by conditional refinement.
\newblock In \emph{Proceedings of the IEEE/CVF conference on computer vision and pattern recognition}, pages 15740--15749, 2021.

\bibitem[Rombach et~al.(2022)Rombach, Blattmann, Lorenz, Esser, and Ommer]{rombach2022high}
Robin Rombach, Andreas Blattmann, Dominik Lorenz, Patrick Esser, and Bj{\"o}rn Ommer.
\newblock High-resolution image synthesis with latent diffusion models.
\newblock In \emph{Proceedings of the IEEE/CVF Conference on Computer Vision and Pattern Recognition}, pages 10684--10695, 2022.

\bibitem[Xu et~al.(2023{\natexlab{b}})Xu, Wang, Cheng, Cao, Shan, Qie, and Gao]{xu2023dream3d}
Jiale Xu, Xintao Wang, Weihao Cheng, Yan-Pei Cao, Ying Shan, Xiaohu Qie, and Shenghua Gao.
\newblock Dream3d: Zero-shot text-to-3d synthesis using 3d shape prior and text-to-image diffusion models.
\newblock In \emph{Proceedings of the IEEE/CVF Conference on Computer Vision and Pattern Recognition}, pages 20908--20918, 2023{\natexlab{b}}.

\bibitem[Luo and Hu(2021)]{luo2021diffusion}
Shitong Luo and Wei Hu.
\newblock Diffusion probabilistic models for 3d point cloud generation.
\newblock In \emph{Proceedings of the IEEE/CVF Conference on Computer Vision and Pattern Recognition}, pages 2837--2845, 2021.

\bibitem[Zhou et~al.(2021)Zhou, Du, and Wu]{zhou20213d}
Linqi Zhou, Yilun Du, and Jiajun Wu.
\newblock 3d shape generation and completion through point-voxel diffusion.
\newblock In \emph{Proceedings of the IEEE/CVF International Conference on Computer Vision}, pages 5826--5835, 2021.

\bibitem[Wu et~al.(2020)Wu, Zhuang, Xu, Zhang, and Chen]{wu2020pq}
Rundi Wu, Yixin Zhuang, Kai Xu, Hao Zhang, and Baoquan Chen.
\newblock Pq-net: A generative part seq2seq network for 3d shapes.
\newblock In \emph{Proceedings of the IEEE/CVF Conference on Computer Vision and Pattern Recognition}, pages 829--838, 2020.

\bibitem[Yang et~al.(2022)Yang, Mo, Lai, Guibas, and Gao]{yang2022dsg}
Jie Yang, Kaichun Mo, Yu-Kun Lai, Leonidas~J Guibas, and Lin Gao.
\newblock Dsg-net: Learning disentangled structure and geometry for 3d shape generation.
\newblock \emph{ACM Transactions on Graphics (TOG)}, 42\penalty0 (1):\penalty0 1--17, 2022.

\bibitem[Zhang et~al.(2023)Zhang, Tang, Niessner, and Wonka]{zhang20233dshape2vecset}
Biao Zhang, Jiapeng Tang, Matthias Niessner, and Peter Wonka.
\newblock 3dshape2vecset: A 3d shape representation for neural fields and generative diffusion models.
\newblock \emph{arXiv preprint arXiv:2301.11445}, 2023.

\bibitem[Chang et~al.(2015)Chang, Funkhouser, Guibas, Hanrahan, Huang, Li, Savarese, Savva, Song, Su, Xiao, Yi, and Yu]{shapenet2015}
Angel~X. Chang, Thomas Funkhouser, Leonidas Guibas, Pat Hanrahan, Qixing Huang, Zimo Li, Silvio Savarese, Manolis Savva, Shuran Song, Hao Su, Jianxiong Xiao, Li~Yi, and Fisher Yu.
\newblock {ShapeNet: An Information-Rich 3D Model Repository}.
\newblock Technical Report arXiv:1512.03012 [cs.GR], Stanford University --- Princeton University --- Toyota Technological Institute at Chicago, 2015.

\bibitem[Koch et~al.(2019)Koch, Matveev, Jiang, Williams, Artemov, Burnaev, Alexa, Zorin, and Panozzo]{Koch_2019_CVPR}
Sebastian Koch, Albert Matveev, Zhongshi Jiang, Francis Williams, Alexey Artemov, Evgeny Burnaev, Marc Alexa, Denis Zorin, and Daniele Panozzo.
\newblock Abc: A big cad model dataset for geometric deep learning.
\newblock In \emph{The IEEE Conference on Computer Vision and Pattern Recognition (CVPR)}, June 2019.

\bibitem[Croitoru et~al.(2023)Croitoru, Hondru, Ionescu, and Shah]{croitoru2023diffusion}
Florinel-Alin Croitoru, Vlad Hondru, Radu~Tudor Ionescu, and Mubarak Shah.
\newblock Diffusion models in vision: A survey.
\newblock \emph{IEEE Transactions on Pattern Analysis and Machine Intelligence}, 2023.

\bibitem[Ho et~al.(2020)Ho, Jain, and Abbeel]{ho2020denoising}
Jonathan Ho, Ajay Jain, and Pieter Abbeel.
\newblock Denoising diffusion probabilistic models.
\newblock \emph{Advances in neural information processing systems}, 33:\penalty0 6840--6851, 2020.

\bibitem[Dhariwal and Nichol(2021)]{dhariwal2021diffusion}
Prafulla Dhariwal and Alexander Nichol.
\newblock Diffusion models beat gans on image synthesis.
\newblock \emph{Advances in neural information processing systems}, 34:\penalty0 8780--8794, 2021.

\bibitem[Nichol and Dhariwal(2021)]{nichol2021improved}
Alexander~Quinn Nichol and Prafulla Dhariwal.
\newblock Improved denoising diffusion probabilistic models.
\newblock In \emph{International Conference on Machine Learning}, pages 8162--8171. PMLR, 2021.

\bibitem[Ho et~al.(2022)Ho, Saharia, Chan, Fleet, Norouzi, and Salimans]{ho2022cascaded}
Jonathan Ho, Chitwan Saharia, William Chan, David~J Fleet, Mohammad Norouzi, and Tim Salimans.
\newblock Cascaded diffusion models for high fidelity image generation.
\newblock \emph{The Journal of Machine Learning Research}, 23\penalty0 (1):\penalty0 2249--2281, 2022.

\bibitem[Kong et~al.(2020)Kong, Ping, Huang, Zhao, and Catanzaro]{kong2020diffwave}
Zhifeng Kong, Wei Ping, Jiaji Huang, Kexin Zhao, and Bryan Catanzaro.
\newblock Diffwave: A versatile diffusion model for audio synthesis.
\newblock \emph{arXiv preprint arXiv:2009.09761}, 2020.

\bibitem[Popov et~al.(2021)Popov, Vovk, Gogoryan, Sadekova, and Kudinov]{popov2021grad}
Vadim Popov, Ivan Vovk, Vladimir Gogoryan, Tasnima Sadekova, and Mikhail Kudinov.
\newblock Grad-tts: A diffusion probabilistic model for text-to-speech.
\newblock In \emph{International Conference on Machine Learning}, pages 8599--8608. PMLR, 2021.

\bibitem[Vahdat et~al.(2021)Vahdat, Kreis, and Kautz]{vahdat2021score}
Arash Vahdat, Karsten Kreis, and Jan Kautz.
\newblock Score-based generative modeling in latent space.
\newblock \emph{Advances in Neural Information Processing Systems}, 34:\penalty0 11287--11302, 2021.

\bibitem[Zeng et~al.(2022)Zeng, Vahdat, Williams, Gojcic, Litany, Fidler, and Kreis]{zeng2022lion}
Xiaohui Zeng, Arash Vahdat, Francis Williams, Zan Gojcic, Or~Litany, Sanja Fidler, and Karsten Kreis.
\newblock Lion: Latent point diffusion models for 3d shape generation.
\newblock \emph{arXiv preprint arXiv:2210.06978}, 2022.

\bibitem[Lyu et~al.(2023)Lyu, Wang, An, Zhang, Lin, and Dai]{lyu2023controllable}
Zhaoyang Lyu, Jinyi Wang, Yuwei An, Ya~Zhang, Dahua Lin, and Bo~Dai.
\newblock Controllable mesh generation through sparse latent point diffusion models.
\newblock In \emph{Proceedings of the IEEE/CVF Conference on Computer Vision and Pattern Recognition}, pages 271--280, 2023.

\bibitem[Nichol et~al.(2022)Nichol, Jun, Dhariwal, Mishkin, and Chen]{nichol2022point}
Alex Nichol, Heewoo Jun, Prafulla Dhariwal, Pamela Mishkin, and Mark Chen.
\newblock Point-e: A system for generating 3d point clouds from complex prompts.
\newblock \emph{arXiv preprint arXiv:2212.08751}, 2022.

\bibitem[Peng et~al.(2021)Peng, Jiang, Liao, Niemeyer, Pollefeys, and Geiger]{peng2021shape}
Songyou Peng, Chiyu Jiang, Yiyi Liao, Michael Niemeyer, Marc Pollefeys, and Andreas Geiger.
\newblock Shape as points: A differentiable poisson solver.
\newblock \emph{Advances in Neural Information Processing Systems}, 34:\penalty0 13032--13044, 2021.

\bibitem[Li et~al.(2023)Li, Duan, Zhou, and Lu]{li2023diffusion}
Muheng Li, Yueqi Duan, Jie Zhou, and Jiwen Lu.
\newblock Diffusion-sdf: Text-to-shape via voxelized diffusion.
\newblock In \emph{Proceedings of the IEEE/CVF Conference on Computer Vision and Pattern Recognition}, pages 12642--12651, 2023.

\bibitem[Chou et~al.(2023)Chou, Bahat, and Heide]{chou2023diffusion}
Gene Chou, Yuval Bahat, and Felix Heide.
\newblock Diffusion-sdf: Conditional generative modeling of signed distance functions.
\newblock In \emph{Proceedings of the IEEE/CVF International Conference on Computer Vision}, pages 2262--2272, 2023.

\bibitem[Shue et~al.(2023)Shue, Chan, Po, Ankner, Wu, and Wetzstein]{shue20233d}
J~Ryan Shue, Eric~Ryan Chan, Ryan Po, Zachary Ankner, Jiajun Wu, and Gordon Wetzstein.
\newblock 3d neural field generation using triplane diffusion.
\newblock In \emph{Proceedings of the IEEE/CVF Conference on Computer Vision and Pattern Recognition}, pages 20875--20886, 2023.

\bibitem[Liu et~al.(2023)Liu, Feng, Black, Nowrouzezahrai, Paull, and Liu]{liu2023meshdiffusion}
Zhen Liu, Yao Feng, Michael~J Black, Derek Nowrouzezahrai, Liam Paull, and Weiyang Liu.
\newblock Meshdiffusion: Score-based generative 3d mesh modeling.
\newblock \emph{arXiv preprint arXiv:2303.08133}, 2023.

\bibitem[Zheng et~al.(2023)Zheng, Pan, Wang, Tong, Liu, and Shum]{zheng2023locally}
Xin-Yang Zheng, Hao Pan, Peng-Shuai Wang, Xin Tong, Yang Liu, and Heung-Yeung Shum.
\newblock Locally attentional sdf diffusion for controllable 3d shape generation.
\newblock \emph{arXiv preprint arXiv:2305.04461}, 2023.

\bibitem[Gupta et~al.(2023)Gupta, Xiong, Nie, Jones, and O{\u{g}}uz]{gupta20233dgen}
Anchit Gupta, Wenhan Xiong, Yixin Nie, Ian Jones, and Barlas O{\u{g}}uz.
\newblock 3dgen: Triplane latent diffusion for textured mesh generation.
\newblock \emph{arXiv preprint arXiv:2303.05371}, 2023.

\bibitem[Nam et~al.(2022)Nam, Khlifi, Rodriguez, Tono, Zhou, and Guerrero]{nam20223d}
Gimin Nam, Mariem Khlifi, Andrew Rodriguez, Alberto Tono, Linqi Zhou, and Paul Guerrero.
\newblock 3d-ldm: Neural implicit 3d shape generation with latent diffusion models.
\newblock \emph{arXiv preprint arXiv:2212.00842}, 2022.

\bibitem[Park et~al.(2019)Park, Florence, Straub, Newcombe, and Lovegrove]{park2019deepsdf}
Jeong~Joon Park, Peter Florence, Julian Straub, Richard Newcombe, and Steven Lovegrove.
\newblock Deepsdf: Learning continuous signed distance functions for shape representation.
\newblock In \emph{Proceedings of the IEEE/CVF conference on computer vision and pattern recognition}, pages 165--174, 2019.

\bibitem[Mescheder et~al.(2019)Mescheder, Oechsle, Niemeyer, Nowozin, and Geiger]{mescheder2019occupancy}
Lars Mescheder, Michael Oechsle, Michael Niemeyer, Sebastian Nowozin, and Andreas Geiger.
\newblock Occupancy networks: Learning 3d reconstruction in function space.
\newblock In \emph{Proceedings of the IEEE/CVF conference on computer vision and pattern recognition}, pages 4460--4470, 2019.

\bibitem[Sharf et~al.(2006)Sharf, Lewiner, Shamir, Kobbelt, and Cohen-Or]{sharf2006competing}
Andrei Sharf, Thomas Lewiner, Ariel Shamir, Leif Kobbelt, and Daniel Cohen-Or.
\newblock Competing fronts for coarse--to--fine surface reconstruction.
\newblock In \emph{Computer Graphics Forum}, volume~25, pages 389--398. Wiley Online Library, 2006.

\bibitem[Huang et~al.(2017)Huang, Zou, Carr, and Ju]{huang2017topology}
Zhiyang Huang, Ming Zou, Nathan Carr, and Tao Ju.
\newblock Topology-controlled reconstruction of multi-labelled domains from cross-sections.
\newblock \emph{ACM Transactions on Graphics (TOG)}, 36\penalty0 (4):\penalty0 1--12, 2017.

\bibitem[Edelsbrunner et~al.(2008)Edelsbrunner, Harer, et~al.]{edelsbrunner2008persistent}
Herbert Edelsbrunner, John Harer, et~al.
\newblock Persistent homology-a survey.
\newblock \emph{Contemporary mathematics}, 453\penalty0 (26):\penalty0 257--282, 2008.

\bibitem[Zhou et~al.(2022)Zhou, Dong, and Lin]{zhou2022learning}
Chi Zhou, Zhetong Dong, and Hongwei Lin.
\newblock Learning persistent homology of 3d point clouds.
\newblock \emph{Computers \& Graphics}, 102:\penalty0 269--279, 2022.

\bibitem[Br{\"u}el-Gabrielsson et~al.(2020)Br{\"u}el-Gabrielsson, Ganapathi-Subramanian, Skraba, and Guibas]{bruel2020topology}
Rickard Br{\"u}el-Gabrielsson, Vignesh Ganapathi-Subramanian, Primoz Skraba, and Leonidas~J Guibas.
\newblock Topology-aware surface reconstruction for point clouds.
\newblock In \emph{Computer Graphics Forum}, volume~39, pages 197--207. Wiley Online Library, 2020.

\bibitem[Mezghanni et~al.(2021)Mezghanni, Boulkenafed, Lieutier, and Ovsjanikov]{mezghanni2021physically}
Mariem Mezghanni, Malika Boulkenafed, Andre Lieutier, and Maks Ovsjanikov.
\newblock Physically-aware generative network for 3d shape modeling.
\newblock In \emph{Proceedings of the IEEE/CVF Conference on Computer Vision and Pattern Recognition}, pages 9330--9341, 2021.

\bibitem[Dong et~al.(2022)Dong, Chen, and Lin]{dong2022topology}
Zhetong Dong, Jinhao Chen, and Hongwei Lin.
\newblock Topology-controllable implicit surface reconstruction based on persistent homology.
\newblock \emph{Computer-Aided Design}, 150:\penalty0 103308, 2022.

\bibitem[Scaramuccia et~al.(2020)Scaramuccia, Iuricich, De~Floriani, and Landi]{scaramuccia2020computing}
Sara Scaramuccia, Federico Iuricich, Leila De~Floriani, and Claudia Landi.
\newblock Computing multiparameter persistent homology through a discrete morse-based approach.
\newblock \emph{Computational Geometry}, 89:\penalty0 101623, 2020.

\bibitem[de~Surrel et~al.(2022)de~Surrel, Hensel, Carri{\`e}re, Lacombe, Ike, Kurihara, Glisse, and Chazal]{de2022ripsnet}
Thibault de~Surrel, Felix Hensel, Mathieu Carri{\`e}re, Th{\'e}o Lacombe, Yuichi Ike, Hiroaki Kurihara, Marc Glisse, and Fr{\'e}d{\'e}ric Chazal.
\newblock Ripsnet: a general architecture for fast and robust estimation of the persistent homology of point clouds.
\newblock In \emph{Topological, Algebraic and Geometric Learning Workshops 2022}, pages 96--106. PMLR, 2022.

\bibitem[Clough et~al.(2020)Clough, Byrne, Oksuz, Zimmer, Schnabel, and King]{clough2020topological}
James~R Clough, Nicholas Byrne, Ilkay Oksuz, Veronika~A Zimmer, Julia~A Schnabel, and Andrew~P King.
\newblock A topological loss function for deep-learning based image segmentation using persistent homology.
\newblock \emph{IEEE transactions on pattern analysis and machine intelligence}, 44\penalty0 (12):\penalty0 8766--8778, 2020.

\bibitem[Otter et~al.(2017)Otter, Porter, Tillmann, Grindrod, and Harrington]{otter2017roadmap}
Nina Otter, Mason~A Porter, Ulrike Tillmann, Peter Grindrod, and Heather~A Harrington.
\newblock A roadmap for the computation of persistent homology.
\newblock \emph{EPJ Data Science}, 6:\penalty0 1--38, 2017.

\bibitem[Allili et~al.(2001)Allili, Mischaikow, and Tannenbaum]{allili2001cubical}
Madjid Allili, Konstantin Mischaikow, and Allen Tannenbaum.
\newblock Cubical homology and the topological classification of 2d and 3d imagery.
\newblock In \emph{Proceedings 2001 international conference on image processing (Cat. No. 01CH37205)}, volume~2, pages 173--176. IEEE, 2001.

\bibitem[Wagner et~al.(2011)Wagner, Chen, and Vu{\c{c}}ini]{wagner2011efficient}
Hubert Wagner, Chao Chen, and Erald Vu{\c{c}}ini.
\newblock Efficient computation of persistent homology for cubical data.
\newblock In \emph{Topological methods in data analysis and visualization II: theory, algorithms, and applications}, pages 91--106. Springer, 2011.

\bibitem[Jaegle et~al.(2021)Jaegle, Gimeno, Brock, Vinyals, Zisserman, and Carreira]{jaegle2021perceiver}
Andrew Jaegle, Felix Gimeno, Andy Brock, Oriol Vinyals, Andrew Zisserman, and Joao Carreira.
\newblock Perceiver: General perception with iterative attention.
\newblock In \emph{International conference on machine learning}, pages 4651--4664. PMLR, 2021.

\bibitem[Bubenik et~al.(2015)]{bubenik2015statistical}
Peter Bubenik et~al.
\newblock Statistical topological data analysis using persistence landscapes.
\newblock \emph{J. Mach. Learn. Res.}, 16\penalty0 (1):\penalty0 77--102, 2015.

\bibitem[Adams et~al.(2017)Adams, Emerson, Kirby, Neville, Peterson, Shipman, Chepushtanova, Hanson, Motta, and Ziegelmeier]{adams2017persistence}
Henry Adams, Tegan Emerson, Michael Kirby, Rachel Neville, Chris Peterson, Patrick Shipman, Sofya Chepushtanova, Eric Hanson, Francis Motta, and Lori Ziegelmeier.
\newblock Persistence images: A stable vector representation of persistent homology.
\newblock \emph{Journal of Machine Learning Research}, 18, 2017.

\bibitem[Karras et~al.(2022)Karras, Aittala, Aila, and Laine]{karras2022elucidating}
Tero Karras, Miika Aittala, Timo Aila, and Samuli Laine.
\newblock Elucidating the design space of diffusion-based generative models.
\newblock \emph{Advances in Neural Information Processing Systems}, 35:\penalty0 26565--26577, 2022.

\bibitem[Hyv{\"a}rinen and Dayan(2005)]{hyvarinen2005estimation}
Aapo Hyv{\"a}rinen and Peter Dayan.
\newblock Estimation of non-normalized statistical models by score matching.
\newblock \emph{Journal of Machine Learning Research}, 6\penalty0 (4), 2005.

\bibitem[Gao et~al.(2022)Gao, Shen, Wang, Chen, Yin, Li, Litany, Gojcic, and Fidler]{gao2022get3d}
Jun Gao, Tianchang Shen, Zian Wang, Wenzheng Chen, Kangxue Yin, Daiqing Li, Or~Litany, Zan Gojcic, and Sanja Fidler.
\newblock Get3d: A generative model of high quality 3d textured shapes learned from images.
\newblock \emph{Advances In Neural Information Processing Systems}, 35:\penalty0 31841--31854, 2022.

\bibitem[Zhang et~al.(2022)Zhang, Nie{\ss}ner, and Wonka]{zhang20223dilg}
Biao Zhang, Matthias Nie{\ss}ner, and Peter Wonka.
\newblock 3dilg: Irregular latent grids for 3d generative modeling.
\newblock \emph{Advances in Neural Information Processing Systems}, 35:\penalty0 21871--21885, 2022.

\bibitem[Wang et~al.(2022)Wang, Liu, and Tong]{wang2022dual}
Peng-Shuai Wang, Yang Liu, and Xin Tong.
\newblock Dual octree graph networks for learning adaptive volumetric shape representations.
\newblock \emph{ACM Transactions on Graphics (TOG)}, 41\penalty0 (4):\penalty0 1--15, 2022.

\bibitem[Maria et~al.(2014)Maria, Boissonnat, Glisse, and Yvinec]{maria2014gudhi}
Cl{\'e}ment Maria, Jean-Daniel Boissonnat, Marc Glisse, and Mariette Yvinec.
\newblock The gudhi library: Simplicial complexes and persistent homology.
\newblock In \emph{Mathematical Software--ICMS 2014: 4th International Congress, Seoul, South Korea, August 5-9, 2014. Proceedings 4}, pages 167--174. Springer, 2014.

\bibitem[Qi et~al.(2017)Qi, Su, Mo, and Guibas]{qi2017pointnet}
Charles~R Qi, Hao Su, Kaichun Mo, and Leonidas~J Guibas.
\newblock Pointnet: Deep learning on point sets for 3d classification and segmentation.
\newblock In \emph{Proceedings of the IEEE conference on computer vision and pattern recognition}, pages 652--660, 2017.

\bibitem[Dosovitskiy et~al.(2020)Dosovitskiy, Beyer, Kolesnikov, Weissenborn, Zhai, Unterthiner, Dehghani, Minderer, Heigold, Gelly, et~al.]{dosovitskiy2020image}
Alexey Dosovitskiy, Lucas Beyer, Alexander Kolesnikov, Dirk Weissenborn, Xiaohua Zhai, Thomas Unterthiner, Mostafa Dehghani, Matthias Minderer, Georg Heigold, Sylvain Gelly, et~al.
\newblock An image is worth 16x16 words: Transformers for image recognition at scale.
\newblock \emph{arXiv preprint arXiv:2010.11929}, 2020.

\end{thebibliography}

\end{document}